\documentclass[lettersize, journal]{IEEEtran}
\usepackage{amsmath, amsfonts}
\usepackage{algorithmic}
\usepackage{array}
\usepackage[caption=false, font=normalsize, labelfont=sf, textfont=sf]{subfig}
\usepackage{textcomp}
\usepackage{stfloats}
\usepackage{url}
\usepackage{verbatim}
\usepackage{graphicx}
\usepackage{tabularx}
\usepackage{pifont}
\usepackage{rotating}
\def\BibTeX{{\rm B\kern-.05em{\sc i\kern-.025em b}\kern-.08em T\kern-.1667em\lower.7ex\hbox{E}\kern-.125emX}}
\usepackage{balance}

\usepackage{tikz}
\usepackage{multirow}
\usepackage{booktabs}
\usepackage{adjustbox}
\usepackage{xcolor}
\usepackage[edges]{forest}
\usepackage{graphicx}
\usepackage{caption}

\usepackage{graphicx}
\usepackage{longtable}

\definecolor{lightgreen}{rgb}{0.56, 0.93, 0.56}
\definecolor{brightlavender}{rgb}{0.75, 0.58, 0.89}
\definecolor{capri}{rgb}{0.0, 0.75, 1.0}
\definecolor{darkpastelgreen}{rgb}{0.01, 0.75, 0.24}

\definecolor{tree-level-1}{RGB}{245,20,85}
\definecolor{tree-level-2}{RGB}{246,86,118}
\definecolor{tree-level-3}{RGB}{248,177,193}
\definecolor{tree-leaf}{RGB}{176,230,198}
\usepackage{dsfont}
\usepackage{booktabs} 
\usepackage{colortbl} 

\definecolor{my_green}{RGB}{51,102,0}
\definecolor{my_red}{RGB}{204, 0, 0}
\newcommand{\cmark}{\textcolor{my_green}{\ding{51}}} 
\newcommand{\xmark}{\textcolor{my_red}{\ding{55}}} 
\usepackage[colorlinks=true, linkcolor=blue, urlcolor=red]{hyperref}
\usepackage{cleveref}

\newcommand{\eg}{\textit{e}.\textit{g}.}
\newcommand{\etal}{\textit{et al}.}

\newcommand{\etc}{\textit{etc}}

\usepackage[square, comma, sort&compress, numbers]{natbib}
\usepackage{url}
\usepackage{fontawesome5} 

\usepackage[edges]{forest}
\usepackage{xcolor}
\usepackage{tikz}

\tikzstyle{my-box}
=[
rectangle,
draw=gray,
rounded
corners,
text
opacity=1,
minimum
height=1.5em,
minimum
width=5em,
inner
sep=2pt,
align=center,
fill
opacity=.5,
line
width=0.8pt,
]
\tikzstyle{leaf}
=[my-box,
minimum
height=1.5em,
fill=pink!10,
text=black,
align=left,font=\normalsize,
inner
xsep=2pt,
inner
ysep=4pt,
line
width=0.8pt,
]

\definecolor{c1}{RGB}{93,191,237} 
\definecolor{c2}{RGB}{237,110,106} 
\definecolor{c3}{RGB}{240,154,69} 
\definecolor{c4}{RGB}{108,222,157} 
\definecolor{c5}{RGB}{205,180,243} 
\definecolor{c6}{RGB}{97,218,184} 
\definecolor{c7}{RGB}{226,115,150}
\definecolor{c8}{RGB}{201,116,201}
\definecolor{c9}{RGB}{23,182,179}
\definecolor{c10}{RGB}{242,157,108}

\newcommand{\benchrow}[1]{\rowcolor{gray!10}\multicolumn{4}{@{}l}{\textbf{#1}}\\}

\begin{document}
    \definecolor{titlecolor}{RGB}{0,0,0}
    \definecolor{hidden-draw}{RGB}{0,0,0}
    \definecolor{lighttealblue}{RGB}{0,0,0}
    \definecolor{lightplum}{RGB}{0, 0, 0}
    \definecolor{harvestgold}{RGB}{0, 0, 0}

    \title{Multimodal Spatial Reasoning in the Large Model Era: A Survey and Benchmarks}
    \author{ 
    $^{\dag}$Xu Zheng$^{1,2}$, 
    $^{\dag}$Zihao Dongfang$^{1}$, 
    $^{*}$Lutao Jiang$^{1}$, 
    $^{*}$Boyuan Zheng$^{1}$, 
    $^{*}$Yulong Guo$^{1}$, 
    Zhenquan Zhang$^{4}$, 
    Giuliano Albanese$^{2}$, 
    Runyi Yang$^{2}$, 
    Mengjiao Ma$^{2}$,
    Zixin Zhang$^{1}$, 
    Chenfei Liao$^{1,5}$,
    Dingcheng Zhen$^{8}$,
    Yuanhuiyi Lyu$^{1}$, \\
    Yuqian Fu$^{2}$,
    Bin Ren$^{6,7}$,
    Linfeng Zhang$^{5}$,
    Danda Paudel$^{2}$, 
    Nicu Sebe$^{7}$, 
    Luc Van Gool$^{2}$, 
    $^\ddag$Xuming Hu$^{1,3}$ \\
    \vspace{2mm}
    \small{
    $^{1}$HKUST(GZ) \quad
    $^{2}$INSAIT, Sofia University “St. Kliment Ohridski” \quad
    $^{3}$HKUST} \quad
    $^{4}$South China University of Technology \\
    \small{
    $^{5}$Shanghai Jiao Tong University \quad 
    $^{6}$University of Pisa \quad 
    $^{7}$University of Trento \quad
    $^{8}$Independent}\\
    \small{$^{\dag}$ Co-first Author; $^{*}$ Core Contributors; $^{\ddag}$ Corresponding Author.}
    \thanks{Manuscript created October, 2020; This work was developed by the
    IEEE Publication Technology Department. This work is distributed under the \LaTeX
    \ Project Public License (LPPL) ( http://www.latex-project.org/ ) version 1.3.
    A copy of the LPPL, version 1.3, is included in the base \LaTeX \ documentation
    of all distributions of \LaTeX \ released 2003/12/01 or later. The opinions expressed
    here are entirely that of the author. No warranty is expressed or implied. User
    assumes all risk.}}


    \twocolumn[{%
    \renewcommand{\twocolumn}[1][]{#1}%
    \maketitle \begin{center}\vspace{-10mm} \includegraphics[width=0.95\linewidth, trim=0 0 0pt 0, clip]{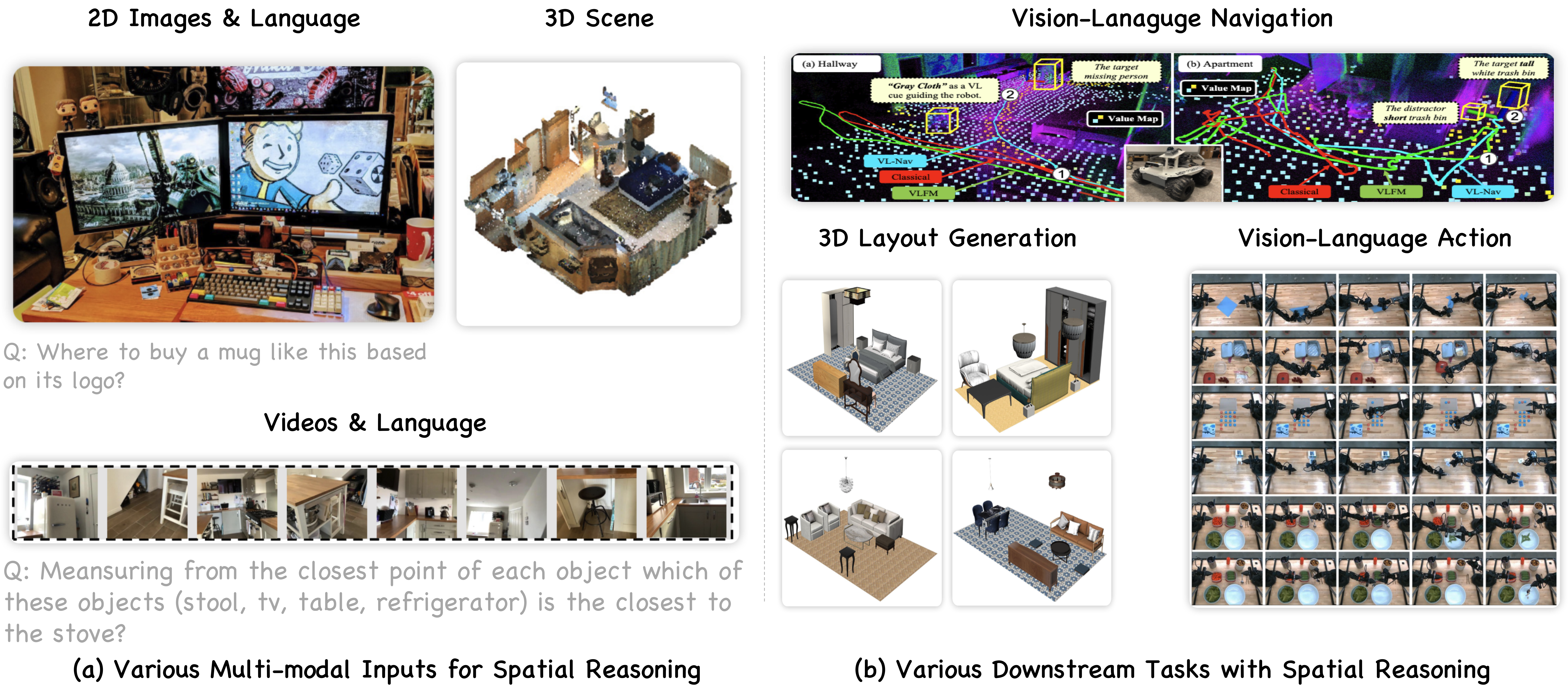}\end{center} \captionof{figure}{ (a) Various multimodal inputs for advanced spatial reasoning with MLLMs, such as 2D images~\cite{su2025pixel}, 3D scenes~\cite{liu20253daxisprompt} and videos~\cite{ouyang2025spacer}. (b) Downstream tasks base or rely on spatial reasoning, such as VLA~\cite{du2025vl}, 3D layout generation~\cite{feng2023layoutgpt}, and vision-language action~\cite{gemini-robotics}. } \label{fig:teaser} }]


\begin{abstract}
Humans possess spatial reasoning abilities that enable them to understand spaces through multimodal observations, such as vision and sound. Large multimodal reasoning models extend these abilities by learning to perceive and reason, showing promising performance across diverse spatial tasks. However, systematic reviews and publicly available benchmarks for these models remain limited.
In this survey, we provide a comprehensive review of multimodal spatial reasoning tasks with large models, categorizing recent progress in multimodal large language models (MLLMs) and introducing open benchmarks for evaluation.
We begin by outlining general spatial reasoning, focusing on post-training techniques, explainability, and architecture. Beyond classical 2D tasks, we examine spatial relationship reasoning, scene and layout understanding, as well as visual question answering and grounding in 3D space.
We also review advances in embodied AI, including vision-language navigation and action models. Additionally, we consider emerging modalities such as audio and egocentric video, which contribute to novel spatial understanding through new sensors.
We believe this survey establishes a solid foundation and offers insights into the growing field of multimodal spatial reasoning.
Updated information about this survey, codes and implementation of the open benchmarks can be found at \url{https://github.com/zhengxuJosh/Awesome-Spatial-Reasoning}.
\end{abstract}




    \begin{figure*}[!pt]
    \centering
    \resizebox{\textwidth}{!}{
        \begin{forest}
            forked edges,
            for tree={
                grow=east,
                reversed=true,
                anchor=base west,
                parent anchor=east,
                child anchor=west,
                base=center,
                font=\large,
                rectangle,
                draw=gray,
                rounded corners,
                align=left,
                text centered,
                minimum width=4em,
                edge+={darkgray, line width=1pt},
                s sep=3pt,
                inner xsep=2pt,
                inner ysep=3pt,
                line width=0.8pt,
                ver/.style={rotate=90, child anchor=north, parent anchor=south, anchor=center},
            },
            where level=1{text width=10em,font=\normalsize,}{},
            where level=2{text width=14em,font=\normalsize,}{},
            where level=3{text width=10em,font=\normalsize,}{},
            where level=4{text width=40em,font=\normalsize,}{},
            where level=5{text width=10em,font=\normalsize,}{},
            [
                \textbf{Multimodal Spatial Reasoning}, ver, line width=0.7mm 
                [
                    \textbf{General MLLM}, fill=c1!60, draw=c1, line width=0mm 
                    [\textbf{Test-Time Scaling}, fill=c1!60, draw=c1, line width=0mm, edge={c1} 
                        [\textbf{Prompt Engineering}, fill=c1!60, draw=c1, edge={c1} 
                            [\textbf{Spatial-MM~\cite{shiri2024empirical}, VSI-Bench~\cite{yang2024thinking}, VoT~\cite{wu2024minds}, \etc} ,leaf, draw=c1, text width=21.5em, edge={c1}] 
                        ]
                        [\textbf{Tool Use}, fill=c1!60, draw=c1, edge={c1} 
                            [\textbf{SpatialScore~\cite{wu2025spatialscore}, SpatialPIN~\cite{ma2024spatialpin}, \etc} ,leaf, draw=c1, text width=21.5em, edge={c1}] 
                        ]
                        [\textbf{Others}, fill=c1!60, draw=c1, edge={c1} 
                            [\textbf{VisuoThink~\cite{wang2025visuothink}, Logic-RAG~\cite{kabir2025logic}, \etc} ,leaf, draw=c1, text width=21.5em, edge={c1}] 
                        ]
                    ]
                    [\textbf{Post-Training}, fill=c1!60, draw=c1, line width=0mm, edge={c1} 
                        [\textbf{SFT}, fill=c1!60, draw=c1, edge={c1} 
                            [\textbf{Multi-SpatialMLLM~\cite{xu2024multispatial}, SpatialVLM~\cite{spatialvlm}, \etc} ,leaf, draw=c1, text width=21.5em, edge={c1}] 
                        ]
                        [\textbf{RL}, fill=c1!60, draw=c1, edge={c1} 
                            [\textbf{Video-R1~\cite{video-r1}, Spatial-R1~\cite{ouyang2025spatial}, \etc} ,leaf, draw=c1, text width=21.5em, edge={c1}] 
                        ]
                    ]
                    [\textbf{Model Design}, fill=c1!60, draw=c1, line width=0mm, edge={c1} 
                            [\textbf{Spatial-MLLM~\cite{wu2025spatialmllm}, SpatialRGPT~\cite{cheng2024spatialrgpt}, Spatial-ORMLLM~\cite{he2025spatialormllm}, \etc},leaf, draw=c1, text width=33em, edge={c1}]
                    ]
                    [\textbf{Explainability}, fill=c1!60, draw=c1, line width=0mm, edge={c1} 
                            [\textbf{Beyond Semantics~\cite{qi2025beyond}, ADAPTVIS~\cite{chen2025why}, RelatiViT~\cite{wen2024transformers}, \etc},leaf, draw=c1, text width=33em, edge={c1}] 
                    ]
                ]
                [
                    \textbf{3D Vision}, fill=c2!60, draw=c2, line width=0mm 
                    [
                        \textbf{3D Visual Grounding}, fill=c2!60, draw=c2, line width=0mm, edge={c2} 
                        [\textbf{3D Input}, fill=c2!60, draw=c2, edge={c2} 
                            [ \textbf{LLM-Grounder~\cite{yang2024llm}, Grounded 3D-LLM~\cite{chen2024grounded}, \etc},leaf, draw=c2, text width=21.5em, edge={c2}] 
                        ]
                        [\textbf{Multi-view Input}, fill=c2!60, draw=c2, edge={c2} 
                            [\textbf{VLM-Grounder~\cite{xu2024vlm}, 3DAxisPrompt~\cite{liu20253daxisprompt}, \etc},leaf, draw=c2, text width=21.5em, edge={c2}] 
                        ]
                        [\textbf{Hybrid of 3D and 2D}, fill=c2!60, draw=c2, edge={c2} 
                            [\textbf{SeeGround~\cite{li2024seeground}, ReasonGrounder~\cite{liu2025reasongrounder}, \etc},leaf, draw=c2, text width=21.5em, edge={c2}] 
                        ]
                    ]
                    [
                        \textbf{3D Scene Reasoning and QA}, fill=c2!60, draw=c2, line width=0mm, edge={c2} 
                        [\textbf{Training-required}, fill=c2!60, draw=c2, edge={c2} 
                            [\textbf{LLaVA-3D~\cite{zhu2024llava}, 3DGraphLLM~\cite{zemskova20243dgraphllm}, \etc},leaf, draw=c2, text width=21.5em, edge={c2}] 
                        ]
                        [\textbf{Training-free}, fill=c2!60, draw=c2, edge={c2} 
                            [\textbf{SpatialPIN~\cite{ma2024spatialpin}, Agent3D-Zero~\cite{zhang2024agent3d}, \etc},leaf, draw=c2, text width=21.5em, edge={c2}] 
                        ]                        
                    ]
                    [
                        \textbf{3D Generation}, fill=c2!60, draw=c2, line width=0mm, edge={c2} 
                        [\textbf{3D Layout Generation}, fill=c2!60, draw=c2, edge={c2} 
                            [\textbf{LayoutGPT~\cite{feng2023layoutgpt}, Layout-your-3D~\cite{zhou2024layout}, \etc},leaf, draw=c2, text width=21.5em, edge={c2}] 
                        ]
                        [\textbf{3DGen as Program}, fill=c2!60, draw=c2, edge={c2} 
                            [\textbf{3D-GPT~\cite{sun20233d}, CAD-Recode~\cite{rukhovich2024cad}, \etc},leaf, draw=c2, text width=21.5em, edge={c2}] 
                        ]                        
                    ]
                ]
                [
                    \textbf{Embodied AI}, fill=c3!60, draw=c3, line width=0mm 
                    [
                        \textbf{Vision-Language Navigation}, fill=c3!60, draw=c3, line width=0mm, edge={c3} 
                        [\textbf{Scene Understanding}, fill=c3!60, draw=c3, edge={c3} 
                            [ \textbf{Spartun3D~\cite{zhang2024spartun3d}, GSA-VLN~\cite{hong2025general},  \etc},leaf, draw=c3, text width=21.5em, edge={c3}] 
                        ]
                        [\textbf{Intention Interpretation }, fill=c3!60, draw=c3, edge={c3} 
                            [ \textbf{AutoSpatial~\cite{kong2025autospatial}, LL3DA~\cite{chen2024ll3da},  \etc},leaf, draw=c3, text width=21.5em, edge={c3}] 
                        ]
                        [\textbf{Planning \& Navigation }, fill=c3!60, draw=c3, edge={c3} 
                            [ \textbf{ NavVLM~\cite{yin2024navigation}, NavCoT~\cite{lin2025navcot},  \etc},leaf, draw=c3, text width=21.5em, edge={c3}] 
                        ]
                    ]
                    [
                        \textbf{Embodied Question Answering}, fill=c3!60, draw=c3, line width=0mm, edge={c3} 
                            [ \textbf{ OpenEQA~\cite{majumdar2024openeqa},  EMBOSR~\cite{hao2024embosr},  \etc},leaf, draw=c3, text width=21.5em, edge={c3}] 
                    ]
                    [
                        \textbf{Embodied Grasping}, fill=c3!60, draw=c3, line width=0mm, edge={c3} 
                            [ \textbf{ ThinkGrasp~\cite{qian2024thinkgrasp}, FreeGrasp~\cite{jiao2025free},  \etc},leaf, draw=c3, text width=21.5em, edge={c3}] 
                    ]
                    [
                        \textbf{Vision-Language Action}, fill=c3!60, draw=c3, line width=0mm, edge={c3} 
                        [ \textbf{ 3D-VLA~\cite{3d-vla}, $\pi0.5$\cite{pi05}, Chat-VLA2 \cite{chat-vla2}, \etc},leaf, draw=c3, text width=21.5em, edge={c3}] 
                    ]
                    [
                        \textbf{Embodied World Model}, fill=c3!60, draw=c3, line width=0mm, edge={c3} 
                            [ \textbf{ TesserAct~\cite{zhen2025tesseract}, EVA~\cite{chi2024eva},  \etc},leaf, draw=c3, text width=21.5em, edge={c3}] 
                    ]
                ]
                [
                    \textbf{Novel Modalities}, fill=c5!60, draw=c5, line width=0mm 
                    [
                        \textbf{Video-based}, fill=c5!60, draw=c5, line width=0mm, edge={c5} 
                            [\textbf{VideoLLaMA2~\cite{damonlpsg2024videollama2}, VideoINSTA~\cite{liao-etal-2024-videoinsta}, Video-R1~\cite{video-r1}, SpaceR~\cite{ouyang2025spacer},~\etc},leaf, draw=c5, text width=33em, edge={c5}] 
                        ]
                        [
                        \textbf{Audio-based}, fill=c5!60, draw=c5, line width=0mm, edge={c5} 
                            [\textbf{STARSS23~\cite{shimada2023starss23}, SpatialSoundQA~\cite{zheng2024bat}, ACORN~\cite{wang2025teaching}, SAVVY~\cite{chen2025savvy}, \etc},leaf, draw=c5, text width=33em, edge={c5}] 
                        ]
                    ]
                ]
                ]
            ]
        \end{forest}
    }
    \caption{Taxonomy for multimodal spatial reasoning with large models.}
    \label{fig:mcot-taxonomy-template}
\end{figure*}
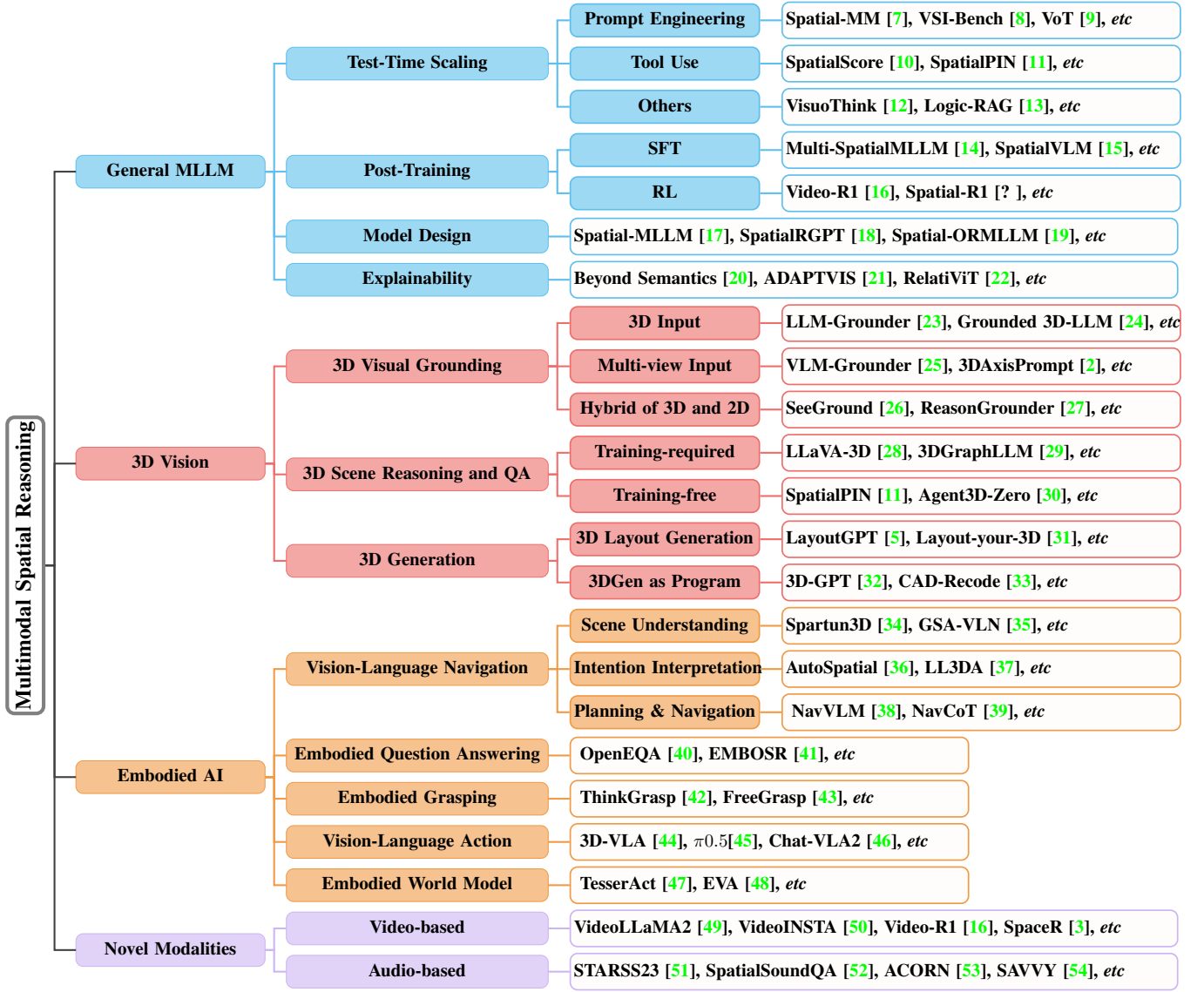

    \section{Introduction}

\subsection{Background}

Spatial reasoning is a fundamental human ability that allows individuals to understand and interact with the world through multimodal inputs, such as vision, sound, and other senses. It supports navigation, comprehension of object relationships, and problem-solving in spatial contexts, as shown in Figure~\ref{fig:teaser}. While large language models (LLMs) have made significant strides in text processing and generation~\cite{bai2023qwen}, their spatial reasoning is limited by their primarily unimodal design~\cite{zha2025enable}. Integrating multimodal information—such as images, audio, and video—into language models offers new opportunities to enhance spatial reasoning, particularly for tasks requiring deep understanding of complex real-world scenarios~\cite{ma2025cityloc, zheng2025mllms, wu2023mars, fu2024objectrelator, li2025scenesplat, brodermann2025cafuser, ma2025scenesplat++}.

Large multimodal reasoning models have emerged as a promising solution, as they are trained to perceive and reason across multiple modalities simultaneously~\cite{lyu2024unibind,zheng2025retrieval,zhou2024eventbind,zheng2024learning,lyu2024omnibind}. These models have shown remarkable performance in a wide range of spatial tasks, from understanding 2D spatial relationships to more complex 3D reasoning. However, despite these advancements, there remains a notable gap in systematically reviewing and evaluating the performance of these emerging models, especially in the context of multimodal spatial reasoning.

\subsection{Contributions}

This survey aims to fill that gap by providing a comprehensive review of the current state of multimodal spatial reasoning with large models, as shown in Figure~\ref{fig:mcot-taxonomy-template}. We begin by reviewing the general landscape of spatial reasoning, focusing on key aspects such as post-training techniques~\cite{spatialvlm,video-r1}, model explainability~\cite{qi2025beyond}, and architecture design~\cite{cheng2024spatialrgpt}. Moving beyond traditional 2D tasks~\cite{wu2025spatialscore}, we delve into more advanced forms of spatial reasoning, including spatial relationship reasoning~\cite{lin2025navcot}, scene and layout understanding~\cite{feng2023layoutgpt}, and grounding visual information in 3D space~\cite{liu2025reasongrounder}.
Furthermore, this paper also explores the intersection of spatial reasoning and embodied AI tasks~\cite{majumdar2024openeqa}, including vision-language navigation and action models~\cite{3d-vla}, where models are required to perform tasks in dynamic environments based on multimodal inputs. We extend the discussion to incorporate the use of emerging modalities such as audio and ego-centric video, which offer distinct opportunities for spatial understanding, particularly in novel sensor environments~\cite{dongfang2025multimodal, zhang2025towards}. In addition to reviewing the existing literature, we introduce open benchmarks for evaluating the performance of MLLMs in spatial reasoning tasks. These benchmarks aim to standardize the evaluation of these models and provide a reliable foundation for future research. The introduction of these benchmarks will also facilitate comparisons across different models and drive advancements in the field by offering standardized testing protocols.

We believe this survey serves as an essential resource for researchers and practitioners in the field of multimodal spatial reasoning, establishing a solid foundation for future work in this critical area. Additionally, we provide access to the codes, implementations, and up-to-date information about the open benchmarks at \url{https://github.com/zhengxuJosh/Awesome-Spatial-Reasnoning}, which can help further advance research in this domain. Through this work, we aim to provide valuable insights into the current challenges and future opportunities in multimodal spatial reasoning with large models, encouraging further exploration and development in this rapidly evolving field.

\begin{table*}[t!]
\centering
\caption{Recent related survey papers on Reasoning in MLLMs.}
\renewcommand{\tabcolsep}{6pt}
\resizebox{\textwidth}{!}{
\begin{tabular}{l|c|l|c}
\toprule
\textbf{Authors} & \textbf{Venue/Date} & \textbf{Main Focus/Analysis} & \textbf{Link} \\ \midrule
Zhou \etal~\cite{zhou2025reinforced} & Arxiv 2025 (May) & RL-based reasoning & \href{https://arxiv.org/pdf/2504.21277}{link} \\
\midrule
Wang \etal~\cite{wang2025short} & Arxiv 2025 (Apr) & Explores small reasoning models, training, inference, and applications & \href{https://arxiv.org/pdf/2504.09100}{link} \\
\midrule
Ke \etal~\cite{ke2025survey} & Arxiv 2025 (Apr) & Discusses inference scaling, learning-to-reason, and agentic systems in LLMs & \href{https://arxiv.org/pdf/2504.09037}{link} \\
\midrule
Zha \etal~\cite{zha2025enable} & Arxiv 2025 (Apr) & Focuses on enabling LLMs with 3D spatial reasoning capabilities & \href{https://arxiv.org/pdf/2504.05786}{link} \\
\midrule
Bi \etal~\cite{bi2025reasoning} & Arxiv 2025 (Apr) & Reviews advancements in multimodal reasoning in LLMs & \href{https://arxiv.org/pdf/2504.03151}{link} \\
\midrule
Chen \etal~\cite{chen2025survey} & Arxiv 2025 (Apr) & Investigates scaling challenges and techniques in LLM reasoning & \href{https://arxiv.org/pdf/2504.02181}{link} \\
\midrule
Chen \etal~\cite{chen2025towards} & Arxiv 2025 (Apr) & Discusses long chain-of-thought approaches for enhancing LLM reasoning & \href{https://arxiv.org/pdf/2503.09567}{link} \\
\midrule
Ali \etal~\cite{forootani2025survey} & Arxiv 2025 (Mar) & Focuses on mathematical reasoning and optimization tasks within LLMs & \href{https://arxiv.org/pdf/2503.17726}{link} \\
\midrule
Wang \etal~\cite{wang2025harnessing} & Arxiv 2025 (Mar) & Reviews efficient reasoning techniques for large-scale LLMs & \href{https://arxiv.org/pdf/2503.24377}{link} \\
\midrule
Plaat \etal~\cite{liu2025efficient} & Arxiv 2025 (Mar) & Explores efficient inference techniques for large reasoning models & \href{https://arxiv.org/pdf/2503.23077}{link} \\
\midrule
Qu \etal~\cite{qu2025survey} & Arxiv 2025 (Mar) & Discusses language and multimodal techniques for efficient reasoning in LLMs & \href{https://arxiv.org/pdf/2503.21614}{link} \\
\midrule
Lin \etal~\cite{lin2025mind} & Arxiv 2025 (Mar) & Focuses on transitioning from language reasoning to multimodal reasoning & \href{https://arxiv.org/pdf/2503.18071}{link} \\
\midrule
Sui \etal~\cite{sui2025stop} & Arxiv 2025 (Mar) & Reviews techniques for reducing inefficiencies in LLM reasoning & \href{https://arxiv.org/pdf/2503.16419}{link} \\
\midrule
Wang \etal~\cite{wang2025multimodal} & Arxiv 2025 (Mar) & Examines the integration of chain-of-thought reasoning with multimodal LLMs & \href{https://arxiv.org/pdf/2503.12605}{link} \\
\midrule
Bandyopadhyay \etal~\cite{bandyopadhyay2025thinking} & Arxiv 2025 (Mar) & Discusses various reasoning strategies implemented in LLMs & \href{https://arxiv.org/pdf/2503.10814}{link} \\
\midrule
Li \etal~\cite{li2025survey} & Arxiv 2025 (Mar) & Focuses on methods to improve causal reasoning abilities in LLMs & \href{https://arxiv.org/pdf/2503.09326}{link} \\
\midrule
Yan \etal~\cite{yan2024survey} & Arxiv 2025 (Feb) & Reviews mathematical reasoning benchmarks and methods in LLMs & \href{https://arxiv.org/pdf/2412.11936}{link} \\
\midrule
Yang \etal~\cite{yang2025code} & Arxiv 2025 (Feb) & Explores code-enhanced reasoning in LLMs, and reasoning-driven code tasks & \href{https://arxiv.org/pdf/2502.19411}{link} \\
\midrule
Li \etal~\cite{li2025system} & Arxiv 2025 (Feb) & Focuses on cognitive reasoning models and LLMs (System 1 vs System 2) & \href{https://arxiv.org/pdf/2502.17419}{link} \\
\midrule
Cheng \etal~\cite{cheng2025empowering} & Arxiv 2025 (Feb) & Discusses integrating logical reasoning in LLMs for more structured outputs & \href{https://arxiv.org/pdf/2502.15652}{link} \\
\midrule
Srivastava \etal~\cite{srivastava2025towards} & Arxiv 2025 (Feb) & Investigates small language models' reasoning abilities and improvements & \href{https://arxiv.org/pdf/2502.11569}{link} \\
\midrule
Xu \etal~\cite{xu2025towards} & Arxiv 2025 (Jan) & Focuses on reinforced reasoning techniques for LLMs & \href{https://arxiv.org/pdf/2501.09686}{link} \\ \midrule
Wang \etal~\cite{wang2024exploring} & Arxiv 2024 (Jan) & Explores the emerging trends and challenges in multimodal reasoning for LLMs & \href{https://arxiv.org/pdf/2401.06805}{link} \\
\midrule
\textbf{\textit{Ours}} & Arxiv 2025 (Aug) & \textbf{\textit{\textcolor{red}{Multimodal spatial reasoning}}} in the large model era & \href{https://github.com/zhengxuJosh/Awesome-Spatial-Reasnoning}{link} \\
\bottomrule
\end{tabular}
}
\label{tab:reasoningsurveyanalysis}
\end{table*}

\subsection{Related Works}

Significant progress has integrated vision, audio, and other modalities with text models, enabling richer spatial reasoning in 2D and 3D. Prior surveys examine related directions but leave gaps relevant to multimodal spatial tasks. For example, Wang \etal~\cite{wang2025short} study small reasoning models but focus on unimodal, low-complexity tasks; Ke \etal~\cite{ke2025survey} analyze inference scaling and agentic systems without deeply addressing multimodal spatial reasoning; and Zha \etal~\cite{zha2025enable} emphasize 3D capabilities but concentrate on implementation details rather than cross-modal evaluation. Broad reviews such as Bi \etal~\cite{bi2025reasoning} summarize multimodal advances but do not propose systematic benchmarks or evaluation frameworks for spatial understanding in dynamic, real-world settings.

Our survey fills this gap by concentrating on \textbf{\textit{multimodal spatial reasoning in the large-model era}}. We categorize spatial tasks (e.g., relationship reasoning, scene understanding, 3D visual grounding), incorporate emerging modalities (audio, egocentric video), and present open benchmarks and evaluation protocols absent from prior work. This focused review aims to provide a concise foundation for advancing research and practical evaluation in multimodal spatial reasoning.

\begin{table}[ht]
\centering
\resizebox{\linewidth}{!}{
\begin{tabular}{l|l}
\midrule
\textbf{Category} & \textbf{Details} \\
\midrule
\textbf{Types} & \begin{tabular}[c]{@{}l@{}}1. \textbf{Localization:} Locate objects in 2D/3D.\\ 2. \textbf{Relation:} Reason about spatial relations.\\ 3. \textbf{Navigation:} Plan paths and optimize actions.\\ 4. \textbf{Pattern:} Detect patterns/symmetries.\\ 5. \textbf{Scaling:} Resize while preserving proportions.\\ 6. \textbf{Transformation:} Apply spatial changes.\\ 7. \textbf{Context:} Interpret positions in context.\\ 8. \textbf{3D Generation:} Synthesize 3D scenes.\\ 9. \textbf{Modeling:} Build scene models for predictions.\\ 10. \textbf{Interaction:} Support real-time spatial interaction. \end{tabular} \\
\midrule
\textbf{Eval} & \begin{tabular}[c]{@{}l@{}}1. \textbf{Multimodal Integration:} Test modality combinations.\\ 2. \textbf{Task Coverage:} VQA, 3D localization, navigation.\\ 3. \textbf{Transparency:} Trace decisions with maps or probes.\\ 4. \textbf{Generalization:} Test adaptability in new environments.\\ 5. \textbf{Embodied Testing:} Measure real-time performance.\\ 6. \textbf{Benchmarking:} Provide reproducible tasks. \end{tabular} \\
\midrule
\textbf{Roadmap} & \begin{tabular}[c]{@{}l@{}}1. \textbf{2D Tasks:} Spatial reasoning in images/videos.\\ 2. \textbf{3D Reasoning:} Grounding, QA, navigation.\\ 3. \textbf{Embodied Reasoning:} Navigation and world models.\\ 4. \textbf{Novel Modalities:} Cross-domain spatial reasoning. \end{tabular} \\
\midrule
\end{tabular}}
\caption{Overview of Spatial Reasoning in MLLMs: Types, Evaluation Protocols, and Roadmap}
\label{tab:spatial_reasoning}
\end{table}

\section{Problem Setup: Multimodal Spatial Reasoning}
\label{sec:problem_setup}

\textbf{Definition.}
Multimodal spatial reasoning aims to infer spatial relations, locations, and actions from heterogeneous inputs and to produce verifiable outputs grounded in space.
Formally, given inputs
$\mathcal{X}=\{x^{\mathrm{img}},x^{\mathrm{vid}},x^{\mathrm{pc}},x^{\mathrm{aud}},x^{\mathrm{text}},\ldots\}$ 
(e.g., RGB images, videos, point clouds, audio, and language) under a specified reference frame (2D/3D/ego/allo), a model predicts
$\mathcal{Y}$ such as \emph{(i)} textual answers/rationales, \emph{(ii)} geometric quantities (boxes, poses, trajectories), or \emph{(iii)} executable actions/plans for embodied settings.
This unifies classic VQA-style queries, 3D grounding, navigation, and layout/scene generation \cite{cheng2024spatialrgpt,wang2024spatial,zhang2024spartun3d,shu2025earthmind,kong2025autospatial}.

\subsection{Types of Spatial Reasoning in MLLMs}

Spatial reasoning in MLLMs spans basic localization to advanced scene modeling. Key types include:
\ding{172} Localization \& Memory: Locate objects in 2D/3D relative to others/observer and track their states over time.
\ding{173} Relation \& Geometry: Reason about spatial relations (above/below/left/right) and metrics (distance, angle, area, volume).
\ding{174} Navigation \& Problem Solving: Plan paths and optimize actions (e.g., shortest routes, spatial puzzles).
\ding{175} Pattern \& Perspective: Detect patterns/symmetries and reason across viewpoints.
\ding{176} Scaling \& Resizing: Model size changes while preserving proportions.
\ding{177} Transformation: Apply rotation, translation, and scaling while maintaining relationships.
\ding{178} Contextualization: Interpret positions under environmental context (e.g., room vs.\ spacecraft).
\ding{179} 3D Model Generation: Synthesize 3D shapes/scenes from spatial cues.
\ding{180} Environmental Modeling: Build scene/world models for prediction and decision making.
\ding{181} Sensing \& Interaction: Support real-time spatial interaction (e.g., AR) via sensors/vision.
These abilities underpin applications from navigation to simulation and interactive systems.

\subsection{Evaluation Protocols for Spatial Reasoning}

Evaluating MLLMs’ spatial reasoning should probe accuracy, robustness, interpretability, and generalization. Key dimensions: \ding{172} Multimodal Integration: Test diverse modality combos (images, text, audio, depth/point clouds, sensors) to assess cross-modal fusion beyond unimodal cues.
\ding{173} Task Coverage: Include VQA, 3D localization, map-based navigation, embodied planning, and scene/image generation to span low- and high-level reasoning.
\ding{174} Process Transparency: Trace decisions via attention maps, intermediate states, or rationale probes to reveal how spatial relations are encoded/manipulated.
\ding{175} Generalization \& Robustness: Evaluate out-of-distribution settings (novel layouts, unseen environments, perturbations) to test adaptability.
\ding{176} Interactive/Embodied Testing: Measure real-time performance for navigation/manipulation and AR/VR, including responsiveness and online updates.
\ding{177} Benchmark Standardization: Provide reproducible suites spanning controlled synthetic tasks and real-world scenarios. Addressing these facets enables comprehensive, comparable assessment of MLLMs’ spatial reasoning and clarifies strengths/weaknesses across applications.

\noindent\textbf{Roadmap.}
We next instantiate this setup across application strata:
(1) \emph{general 2D image/video tasks with MLLMs}, 
(2) \emph{3D spatial reasoning} (grounding, QA, navigation), and 
(2) \emph{embodied spatial reasoning} (VLN, VLA, world model), and 
(3) \emph{novel modalities \& cross-domain settings}.
Each section maps back to the taxonomy above and adopts the evaluation dimensions outlined here.

    \begin{figure}[t!]
  \centering
\includegraphics[width=.95\linewidth]{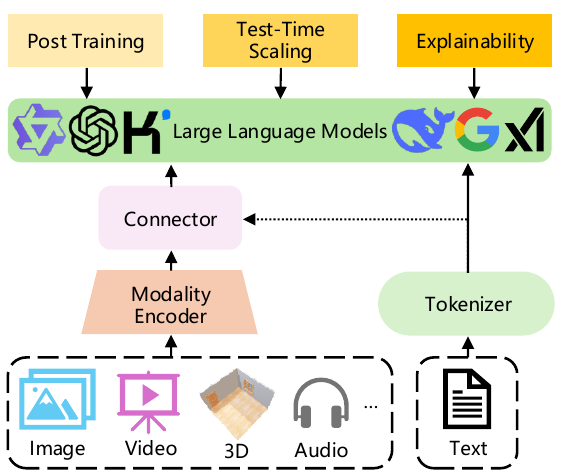}
  \caption{Typical MLLM architecture and strategies.}
  \label{fig:general_arch}
\end{figure}

\section{General Multimodal Spatial Reasoning}

General multimodal spatial reasoning refers to MLLMs’ ability to understand and reason about spatial relationships across visual and textual inputs. It encompasses tasks such as visual question answering (VQA) on spatial relations, object localization, perspective understanding, 3D comprehension, and navigation. These tasks require aligning visual perception with linguistic expressions of spatial concepts like “above,” “behind,” and “to the left of.”
As shown in Figure \ref{fig:general_arch}, current research enhances spatial reasoning in multimodal models along four main directions:
\ding{172} Test-time scaling to boost inference-time capability;
\ding{173} Post-training methods such as supervised fine-tuning and reinforcement learning on spatial datasets;
\ding{174} Architectural improvements for richer spatial encoding; and
\ding{175} Explainability studies to reveal limitations and failure modes in spatial reasoning.

\subsection{Test-Time Scaling Methods}

Test-time scaling methods offer training-free strategies to enhance MLLMs’ spatial reasoning during inference. Instead of retraining or fine-tuning, these approaches leverage improved prompting, tool-assisted reasoning, and external modality integration. Existing works can be broadly grouped into three categories based on their methodological focus.

\begin{table}[t]
\centering
\caption{Comparison of prompt engineering methods for multimodal spatial reasoning. We summarize key ideas and prompt types of representative approaches.}
\label{tab:prompt_engineering_comparison}
\renewcommand{\arraystretch}{1.2}
\setlength{\tabcolsep}{2pt}
\resizebox{\linewidth}{!}{
\begin{tabular}{l|c|p{4cm}}
\toprule
\textbf{Method} & \textbf{Prompt Type} & \textbf{Key Idea / Mechanism} \\
\midrule
\textbf{TopViewRS}~\cite{li2024topviewrs} & Textual & Uses simple top-view templates as baseline prompts for spatial reasoning. \\
\midrule
\textbf{VSI-Bench}~\cite{yang2024thinking} & Textual (graph-structured) & Guides models to build and use cognitive graphs for spatial distance reasoning. \\
\midrule
\textbf{OmniSpatial}~\cite{omnispatial} & Textual (CoT) & Applies Chain-of-Thought reasoning to spatial VQA tasks. \\
\midrule
\textbf{SCABenchmark}~\cite{wu2025scog} & Textual (structured cues) & Adds coordinates and reference frames to improve spatial understanding. \\
\midrule
\textbf{Spatial-MM}~\cite{shiri2024empirical} & Visual & Uses bounding boxes or scene graphs to enhance spatial reasoning accuracy. \\
\midrule
\textbf{Mind’s Eye (VoT)}~\cite{wu2024minds} & Visual / Hybrid & Visualizes reasoning steps as spatial traces to aid understanding. \\
\midrule
\textbf{SpatialPIN}~\cite{ma2024spatialpin} & Progressive & Decomposes complex spatial queries into multi-stage sub-tasks. \\
\midrule
\textbf{SpatialPrompt}~\cite{liao2024reasoning} & Quantitative / Textual & Establishes spatial anchors for stepwise geometric reasoning. \\
\midrule
\textbf{SpatialMind}~\cite{zhang2024spatial} & Structured / Multi-modal & Integrates scene representations with task-specific reasoning plans. \\
\bottomrule
\end{tabular}
}
\end{table}

\subsubsection{Prompt Engineering}
Prompt engineering is the most direct and lightweight approach to enhance spatial reasoning in MLLMs without external tools or fine-tuning. Recent work explores how carefully crafted prompts can better elicit models’ latent spatial reasoning abilities. Although Chain-of-Thought (CoT) prompting has achieved notable success in general reasoning, its direct application to spatial tasks yields limited gains. To address this, researchers have proposed specialized prompting strategies tailored for spatial understanding, as shown in Table~\ref{tab:prompt_engineering_comparison}.

Early methods, such as TopViewRS~\cite{li2024topviewrs}, introduce simple templates but show only marginal improvements. VSI-Bench~\cite{yang2024thinking} demonstrates that explicitly instructing MLLMs to build cognitive graphs enhances spatial question answering, whereas standard CoT fails. Similarly, OmniSpatial~\cite{omnispatial} finds textual CoT ineffective for complex perspective-taking. SCABenchmark~\cite{wu2025scog} further analyzes prompt formats and frames of reference, showing that explicit geometric and relational cues—like coordinates and reference frames—outperform long, free-form CoT reasoning. Beyond text, visual prompting has proven complementary. Spatial-MM~\cite{shiri2024empirical} shows that supplying bounding boxes or scene graphs—either annotated or self-generated—greatly improves multi-hop spatial reasoning, where CoT alone fails. Mind’s Eye~\cite{wu2024minds} extends this with the Visualization-of-Thought paradigm, where the model visualizes reasoning traces during inference, significantly boosting 2D spatial reasoning accuracy.

Additionally, progressive prompting frameworks decompose complex queries into manageable steps. SpatialPIN~\cite{ma2024spatialpin} employs multi-stage prompting with dense visual priors from multiple vision foundation models, demonstrating the benefits of structured, incremental reasoning. For quantitative spatial reasoning, SpatialPrompt~\cite{liao2024reasoning} improves performance by establishing explicit spatial anchors and prompting stepwise transformations relative to them. SpatialMind~\cite{zhang2024spatial} integrates scene representations—modeled as object-centric text, 2D grids, or 3D maps—with question-type–specific reasoning plans (e.g., locate → transform → compare), guiding more systematic inference at test time.

\noindent \textit{\textbf{Insights \& Discussion.}}
The evolution from simple CoT prompting to spatially structured prompting reveals a key distinction between linguistic and spatial reasoning in MLLMs. While textual CoT assumes that verbalizing intermediate steps improves reasoning, spatial reasoning requires explicit modeling of visual relations—through visual traces, structured graphs, or reference-based transformations. This indicates that effective spatial prompting depends less on longer reasoning chains and more on aligning prompt representations with the inherently visual and relational nature of spatial cognition. \textbf{\textit{Future work}} may explore adaptive prompting frameworks that automatically select the most suitable representational format—textual, visual, or hybrid—based on the type of spatial query and reasoning context.

\begin{table}[t!]
\centering
\caption{Summary of tool-usage methods for multimodal spatial reasoning. $\checkmark$ indicates the method supports the feature.}
\label{tab:tool_usage_summary}
\setlength{\tabcolsep}{5pt}
\resizebox{\linewidth}{!}{
\begin{tabular}{l|ccccccccc}
\toprule
\textbf{Method} &
\rotatebox{90}{UI ops (crop/seg)} &
\rotatebox{90}{2D perception} &
\rotatebox{90}{3D recon} &
\rotatebox{90}{Append images/ traces} &
\rotatebox{90}{Serialize tokens/ BEV} &
\rotatebox{90}{Render novel views} &
\rotatebox{90}{Agentic control} &
\rotatebox{90}{Plan–Execute} &
\rotatebox{90}{ReAct} \\
\midrule
IoT~\cite{zhou2024image}                 & \cmark & \xmark & \xmark & \cmark & \xmark & \xmark & \xmark & \xmark & \xmark \\
Struct2D~\cite{zhu2025struct2d}          & \xmark & \cmark & \xmark & \xmark & \cmark & \xmark & \xmark & \xmark & \xmark \\
Lee \textit{et al.}~\cite{lee2024perspective} & \xmark & \cmark & \xmark & \xmark & \cmark & \xmark & \xmark & \xmark & \xmark \\
ZeroVLM~\cite{meng2024know,liu2023zero}  & \xmark & \xmark & \cmark & \cmark & \xmark & \cmark & \xmark & \xmark & \xmark \\
SpatialPIN~\cite{ma2024spatialpin}       & \xmark & \cmark & \cmark & \xmark & \cmark & \cmark & \xmark & \xmark & \xmark \\
VADAR~\cite{marsili2025visual}           & \xmark & \cmark & \xmark & \xmark & \cmark & \xmark & \cmark & \cmark & \xmark \\
SpatialAgent~\cite{wu2025spatialscore}   & \xmark & \cmark & \xmark & \xmark & \cmark & \xmark & \cmark & \cmark & \cmark \\
\bottomrule
\end{tabular}
}
\end{table}

\subsubsection{Tool Usage}
Integrating tools at test time enhances MLLMs’ spatial reasoning by providing explicit geometric or structural priors without modifying the base model. As in Table~\ref{tab:tool_usage_summary}, three main tool families have emerged. First, \textbf{\textit{UI-style visual operations}} (\eg, crop, zoom, mark, edge, and segmentation) expose fine-grained spatial cues often missed by MLLMs. For instance, Image-of-Thought~\cite{zhou2024image} directs the model to plan and execute short visual operation sequences, generating “visual rationales” that are fed back alongside text reasoning. Second, \textbf{\textit{2D perception modules}}—such as object detection, orientation, depth, and pose estimation—convert pixels into structured, object-centric facts. Struct2D~\cite{zhu2025struct2d} renders a BEV canvas with filtered object marks and metadata (IDs, categories, coordinates), while Lee et al.~\cite{lee2024perspective} constructed abstract scene layouts to support perspective transformations. Third, \textbf{\textit{3D reconstruction tools}} lift images into view-consistent geometry for perspective-based reasoning. ZeroVLM~\cite{meng2024know} employs Zero-1-to-3~\cite{liu2023zero} to synthesize novel views and pair them with “view prompts” that anchor camera relations, and SpatialPIN~\cite{ma2024spatialpin} partially reconstructs lightweight 3D objects for downstream spatial queries.

Given these tools, inference-time integration generally follows three escalating patterns. First, some methods append tool-generated images or traces to the input: IoT concatenates cropped or segmented snippets as visual evidence, while ZeroVLM stitches multi-view mosaics with view-aware prompts~\cite{zhou2024image, meng2024know}.
Second, others serialize perception into structured tokens or sketches: Struct2D supplies a BEV bitmap with concise object metadata, and Lee et al. inject numeric orientations and perspective descriptors to convert allocentric queries into egocentric ones~\cite{zhu2025struct2d, lee2024perspective}.
Finally, 3D-aware approaches render novel views from reconstructed geometry: ZeroVLM generates left/right/random perspectives to test viewpoint sensitivity, while SpatialPIN’s partial 3D lifting enables virtual viewpoints that re-ground spatial relations~\cite{meng2024know, ma2024spatialpin}.

Beyond single-shot prompting, modern systems increasingly control tools through agentic policies at inference. VADAR (Visual Agentic AI)~\cite{marsili2025visual} designs a dynamic API and synthesizes short programs that call specialized modules (detector, depth, pose) on demand—illustrating “plan-to-execute” tool use via code generation for reliable multi-step reasoning. SpatialScore’s SpatialAgent~\cite{wu2025spatialscore} provides a standardized multi-agent framework with nine spatial tools and two control paradigms: a hierarchical Plan–Execute pipeline and an interleaved ReAct mode that alternates reasoning and action, enabling consistent cross-method evaluation. Diagnostics further reveal which tool outputs matter most. Ravi~\etal~\cite{ravi2024out} show in Disjoint-3DQA that trajectories or BEV features offer limited gains across non-co-visible frames, while oracle 3D coordinates yield substantial improvements—highlighting metrically faithful 3D states or persistent scene memory as the most effective feedback signals.

\noindent \textit{\textbf{Insights \& Discussion.}} 
Test-time tool use works by externalizing geometry into inputs MLLMs already consume—visual traces, structured tokens, and novel views—rather than elongating textual CoT. Gains are largest when signals are metrically grounded (poses, coordinates, calibrated depth) and agentic controllers compose tools into reusable subroutines, improving perspective shifts, occlusions, and multi-object relations without retraining. \textbf{\textit{\ding{172} Remaining issues}}: perception and view-synthesis errors propagate without uncertainty handling; 2D proxies (BEV, trajectories) poorly approximate metric 3D state; temporal persistence is weak—no durable, object-centric world memory; and tool outputs lack standardized units/frames, harming alignment and reproducibility. Multi-tool pipelines also add cost and latency for open-world, long-horizon tasks. \textbf{\textit{\ding{173} Promising directions}}: maintain a persistent object-centric scene memory with cross-view/time checks and lightweight geometric self-verification; standardize tool outputs (schemas for objects/cameras/constraints with calibrated uncertainty) to enable evidence weighting and conflict resolution; and develop budget-aware controllers that switch between Plan–Execute and ReAct, add verify–reflect loops, and distill heavy chains into compact prompts/plugins—evaluated with utility–cost–robustness metrics in long-horizon, non-co-visible, open-world regimes.

\subsubsection{Others}
Beyond prompt engineering and tool use, several \textit{training-free} inference strategies improve spatial reasoning. The first category is self-consistency voting.
Sample multiple reasoning chains and take a \emph{consensus} to stabilize answers under perspective shifts and multi-object relations.
Secondly, multimodal search explores and prunes visual-spatial reasoning paths at test time; \eg, \textsc{VisuoThink} performs \emph{look-ahead tree search} over interleaved visual-textual steps and selects the best-scoring solution under spatial constraints~\cite{wang2025visuothink}.
There are also retrieval-augmented generation (RAG) methods. Inject external spatial knowledge at inference. \textsc{Logic-RAG}~\cite{kabir2025logic} builds a dynamic first-order logic knowledge base (object positions/relations) from visual input and feeds these facts to the model, increasing driving-scene spatial accuracy from \(\sim\)55–75\% to \(>\)80--90\%. Grounding in retrieved maps/KBs or computed facts reduces hallucinations and sharpens spatial relations.

\noindent \textit{\textbf{Insights \& Discussion.}}
Enhancing spatial reasoning in MLLMs often requires more than static prompts or single-pass outputs. Exploring multiple reasoning paths, retrieving external spatial knowledge, performing light test-time adaptation, and preserving spatial context collectively scale inference-time capability and complement prompt/tool methods. These approaches carry trade-offs—\eg, multi-sampling and adaptation increase compute, while retrieval depends on knowledge quality—but they point toward MLLMs that dynamically and reliably reason about space with higher accuracy.

\subsection{Post-Training Methods}

Post-Training methods enhance spatial reasoning by adapting MLLMs after pre-training, mainly through supervised fine-tuning and reinforcement learning (RL). 
These approaches rely on spatially targeted datasets, rewards, and curricula to strengthen model understanding of geometry and motion.

\subsubsection{Supervised Fine-tuning (SFT)}
SFT advances spatial reasoning by progressively broadening supervision from domain-specific static scenes to dynamic, temporally grounded reasoning. 
On the data side, domain-grounded QA continues to seed robust priors. \textsc{CityGPT}~\cite{citygpt} injects urban navigation and landmark knowledge through structured instructions, while \textsc{Multi-SpatialMLLM}~\cite{xu2024multispatial} moves from single images to multi-frame settings, annotating frame-level relations (\eg, depth, camera/object motion) to capture persistence and occlusion. Extending this trend, \textsc{LLaVA-ST}~\cite{st-align} aligns fine-grained spatio-temporal understanding by coupling language with explicit coordinates and temporal anchors, and \textsc{ST-Think}~\cite{ego-st} focuses the lens on egocentric 4D reasoning to expose viewpoint changes and long-horizon temporal cues missing from static corpora. Synthetic pipelines complement real data: \textsc{SAT}~\cite{SAT} generates interactive, motion-centric tasks in simulation to cover self-motion and object-motion factors, and \textsc{SpaRE}~\cite{ogezi2025spare} automatically distills spatial QA from long-form descriptions to relieve the long-tail sparsity of rare relations. In between these regimes, \textsc{SpatialVLM}~\cite{spatialvlm} augments instruction tuning with region tags and relative-position tokens (left-of, in-front-of, between, \etc), pairing layout-driven QA and referring expressions so that textual predicates are explicitly bound to coordinates/regions rather than inferred implicitly.

Training strategy then ties these sources together. Curricula that progress from perception to composition remain effective: \textsc{Sparkle}~\cite{tang2024sparkle} stages supervision from detection/localization toward multi-hop spatial reasoning. In parallel, motion-aware instruction tuning such as \textsc{ST-VLM}~\cite{ko2024stvlm} makes kinematics explicit with trajectory-style hints. Multi-stage alignment further stabilizes learning: \textsc{LLaVA-ST}~\cite{st-align} couples semantic-to-coordinate alignment with video-aware objectives, whereas \textsc{SAT}~\cite{SAT} interleaves dynamic spatial tasks as higher-level “sub-curricula” to encourage transfer from static to viewpoint-shifting scenarios. Looking beyond plain instruction tuning, ``thinking'' overlays also matter: \textsc{Visualization-of-Thought}~\cite{imagine} and \textsc{visual+textual thinking}~\cite{liang2025vts} introduce multimodal reasoning traces (textual steps with region/coordinate cues), nudging the model to externalize intermediate spatial inferences rather than collapsing them into a single answer token.


\noindent \textit{\textbf{Insights \& Discussion.}}  
SFT highlights the value of task-specific data and structured curricula for strengthening spatial reasoning in MLLMs. Compared with pre-training alone, spatially grounded supervision enables models to internalize explicit spatial relations, motion cues, and temporal dependencies often missing in general multimodal data. Methodologically, SFT studies show that gradual exposure to increasing spatial complexity—starting from low-level perception (\eg, object localization) to higher-order reasoning (\eg, trajectory prediction, multi-hop inference)—consistently improves model performance. Incorporating temporally annotated or motion-aware datasets further allows models to reason over both static configurations and dynamic evolution. Nonetheless, current SFT methods depend heavily on human-labeled or synthetic data, limiting scale and diversity. \textbf{\textit{Future work}} could focus on automatically generating spatial annotations, leveraging self-supervised pretexts, or designing adaptive multi-task curricula that balance static and dynamic reasoning. Ultimately, effective SFT should align supervision with the cognitive structure of spatial reasoning, bridging perception and high-level spatial understanding.

\begin{table}[t]
\centering
\caption{Comparison of reinforcement learning methods for spatial reasoning in MLLMs. $\checkmark$ indicates the presence of a feature, $\times$ indicates absence.}
\label{tab:rl_comparison_rotated}
\renewcommand{\arraystretch}{1.2}
\setlength{\tabcolsep}{10pt}
\resizebox{\linewidth}{!}{
\begin{tabular}{l|cccccc}
\toprule
\textbf{Method} &
\rotatebox{90}{Reward Design} &
\rotatebox{90}{Process-level Reward} &
\rotatebox{90}{Curriculum Learning} &
\rotatebox{90}{Self-Play / Exploration} &
\rotatebox{90}{3D Spatial Metrics} &
\rotatebox{90}{Temporal Consistency} \\
\midrule
\textbf{Video-R1}~\cite{video-r1} & \checkmark & \xmark & \xmark & \xmark & \xmark & \checkmark \\
\textbf{Spatial-R1}~\cite{ouyang2025spacer} & \checkmark & \checkmark & \xmark & \xmark & \xmark & \checkmark \\
\textbf{MetaSpatial}~\cite{pan2025metaspatial} & \checkmark & \checkmark & \checkmark & \xmark & \checkmark & \xmark \\
\textbf{R1-Zero}~\cite{vis100k} & \checkmark & \xmark & \xmark & \checkmark & \xmark & \xmark \\
\textbf{ST-Think}~\cite{ego-st} & \checkmark & \xmark & \xmark & \xmark & \xmark & \checkmark \\
\textbf{M2-Reasoning}~\cite{wang2025m2reasoning} & \checkmark & \xmark & \checkmark & \xmark & \xmark & \checkmark \\
\bottomrule
\end{tabular}
}
\end{table}

\subsubsection{Reinforcement Learning (RL)}  
RL enhances spatial reasoning by optimizing models through reward-driven feedback rather than explicit supervision.

On rewards, as in Table~\ref{tab:rl_comparison_rotated} \textsc{Video-R1}~\cite{video-r1} introduces a time-order–aware signal (\eg, preferring correct answers on ordered vs.\ shuffled clips) to explicitly reward temporal use, while \textsc{Spatial-R1}/\textsc{SpaceR}~\cite{ouyang2025spacer} extends beyond outcome rewards to process-aware credit for intermediate steps (\eg, partial route/landmark correctness, local relation checks) to improve reward stability. For 3D layout and interaction, \textsc{MetaSpatial}~\cite{pan2025metaspatial} blends format checks, physical feasibility, and rendering-based validation—together with object-level modulation—for consistent spatial plans. Unifying general and spatial reasoning, \textsc{M2-Reasoning}~\cite{wang2025m2reasoning} adopts task-specific RLVR signals (\eg, coordinate/ordering correctness) so that spatial subtasks contribute targeted feedback without derailing broader multimodal skills.

Training strategies typically follow a staged recipe—\emph{warm up with SFT, then refine with RL, and finally stabilize with self-improvement}. \textsc{Video-R1}~\cite{video-r1} uses SFT to initialize video reasoning and then applies temporally sensitive RL to consolidate it. Similarly, \textsc{ST-Think}~\cite{ego-st} employs Long-CoT SFT followed by GRPO; meanwhile, reverse thinking is used as the explicit thought style in RL, strengthening bidirectional spatial recall. \textsc{MetaSpatial}~\cite{pan2025metaspatial} employs curriculum-style increases in scene difficulty and multi-round refinement so that rewards stay informative as tasks grow more complex. Self-play closes the loop: R1-Zero–like training~\cite{vis100k} generates and solves spatial puzzles autonomously, reducing dependence on human labels and converting search over solutions into search over training data. In broader multi-task settings, \textsc{M2-Reasoning}~\cite{wang2025m2reasoning} interleaves spatial RLVR with general-purpose tasks and dynamic scheduling, mitigating interference while retaining cross-task transfer.



Overall, these approaches illustrate how RL advances spatial reasoning from two complementary angles: (1) \emph{reward design}, which explicitly encodes geometric and temporal correctness; and (2) \emph{self-improvement}, where models iteratively refine reasoning through autonomous exploration. Compared with supervised fine-tuning, RL offers a more flexible framework for post-training adaptation—enhancing spatial consistency, dynamic reasoning, and generalization without modifying the base architecture.

\noindent \textit{\textbf{Insights \& Discussion.}}  
Reinforcement learning (RL) provides a powerful framework for improving spatial reasoning in MLLMs by optimizing beyond static supervision. The reviewed methods reveal a clear evolution: from composite task-level rewards (\textsc{Video-R1}) to process-level and curriculum-based optimization (\textsc{Spatial-R1}, \textsc{MetaSpatial}), and finally to autonomous self-play learning (\textsc{R1-Zero}). This progression reflects a shift from externally guided training toward self-improving spatial cognition.  

Two primary insights emerge. First, \emph{reward granularity} matters—integrating intermediate reasoning rewards and geometric correctness encourages stable and interpretable spatial learning. Second, \emph{autonomous exploration} enables continual improvement without reliance on labeled data, a promising direction for scalable spatial intelligence.  

However, current RL frameworks remain constrained by high computational cost, reward sparsity, and limited generalization across 2D–3D–temporal domains. Future research could develop hybrid paradigms that combine RL with supervised fine-tuning or self-distillation, using automatically generated spatial feedback signals. Advancing toward richer, self-supervised spatial rewards and cross-domain generalization will be key to achieving more human-like spatial reasoning in multimodal large language models.

\subsection{MLLM Architectural Modifications}
Beyond post-training, architectural changes are essential for enabling MLLMs to reason about space effectively. Most MLLMs adopt a standard three-part structure—a pre-trained LLM, a visual encoder, and a modality alignment interface~\cite{radford2021learning,cherti2023reproducible,sun2023eva,tschannen2025siglip,zhai2023sigmoid,lyu2024unibind}. However, spatial reasoning demands explicit preservation of positional and geometric information, which these components alone cannot ensure. Recent studies have thus proposed modifications to inject spatial knowledge either at the input level or via specialized model components.

\subsubsection{Enhancing Input Representations}  
One strategy is to augment the model inputs with additional spatial cues so that the LLM can infer geometric relations without changing the core architecture. 

The most straightforward one, \textsc{SpatialLLM}~\cite{ma2025spatialllm}, adopts a composite 3D information design, where the vision front-end mixes features from a language-supervised encoder (CLIP) with features from a self-supervised encoder (DINOv2 or MAE) to improve the 3D perception capability at the input level.
Going further, \textsc{MPDrive}~\cite{zhang2025mpdrive} adds an extra ``marker'' channel to each video frame, overlaying simple glyphs or numeric labels at detected object centers. The model processes the original RGB frame and this marker map in parallel (dual-stream), effectively bridging visual coordinates with language; this yields improved spatial understanding on autonomous driving VQA tasks. 
Similarly, \textsc{LocVLM}~\cite{ranasinghe2024learning} appends normalized $(x,y)$ location coordinates of salient objects directly into the text prompt (treating location as part of the language input). By doing so, the LLM is encouraged to reason about spatial relations (\eg, ``left of'', ``inside of'') using these coordinate tokens, all without altering the pre-trained vision encoder or adding new visual branches. Both methods inject explicit spatial information into the model’s context, which in turn guides the language model to produce spatially-aware descriptions and answers.
Another direction is to incorporate depth and 3D cues as part of the input. \textsc{SpatialBot}~\cite{cai2024spatialbot} feeds the model with both an RGB image and its corresponding depth image (\eg, from a monocular depth estimator), essentially giving the MLLM a pseudo-3D view of the scene. This simple input-level fusion of color and depth significantly boosts the model’s depth perception and spatial QA performance, as evidenced by improvements on the SpatialQA benchmark and embodied AI tasks. Rather than images, \textsc{SSR}~\cite{liu2025ssr} leverages depth information in textual form: it converts raw depth maps into structured natural-language rationales describing the 3D layout (\eg, relative distances, sizes, and occlusions). These intermediate text descriptions are provided to the LLM (as a chain-of-thought prompt) to guide its reasoning and are later distilled into latent embeddings for efficiency. This rationale-guided approach enables the model to utilize depth cues for higher-order spatial reasoning without requiring special sensors at inference. In a similar vein, other works enrich the model’s visual context by supplying multiple views or an explicit 3D scene representation. For instance, the \textsc{Spatio-Temporal LLM} framework~\cite{zheng2025stllm} can input an entire point cloud of the environment alongside an egocentric video clip, allowing the LLM to consider the global 3D scene while also tracking temporal events. Experiments show that feeding both the holistic point cloud and video frames (plus text) enables better spatial understanding of environments and improves temporal grounding of actions. Likewise, \textsc{MM-Spatial}~\cite{daxberger2025mmspatial} explores training MLLMs with multi-view images of a scene and their associated metric depth values. By exposing the model to multiple perspectives and precise depth measurements during fine-tuning (via the CA-VQA dataset), MM-Spatial achieves state-of-the-art 3D spatial understanding; notably, it can estimate object sizes and distances with accuracy on par with dedicated monocular depth estimators. 
In summary, these input-centric approaches enhance spatial reasoning by explicitly encoding geometry into the model’s inputs (either as augmented images or as location/depth tokens in text). This mitigates the loss of spatial information in standard vision backbones and provides the LLM with a richer basis for spatial inference.

\noindent \textit{\textbf{Insights \& Discussion.}}
Input-centric augmentation remains minimally invasive: marker channels or coordinate tokens guide the LLM toward geometry without altering backbones, while depth, multi-view, or point-cloud evidence supplies 3D context that strengthens grounding. Yet performance is tightly coupled to detector/depth fidelity, and longer contexts strain alignment and attention memory. Uncertainty-aware spatial tokenizers and differentiable 2D–3D projectors that compress geometry, paired with curricula that progress from single-view to spatio-temporal inputs, are likely to curb shortcut reliance and improve cross-domain generalization.

\subsubsection{Redesigning Spatial Reasoning Modules}  
An alternative (and complementary) approach introduces dedicated architectural modules that are tailored for spatial and relational reasoning. Here, the base MLLM architecture is extended with new components (or entire sub-networks) that preserve spatial structure through the model’s internal representations.
For example, \textsc{Spatial-MLLM}~\cite{wu2025spatialmllm} introduces a dedicated \emph{spatial encoder} built on a lightweight VGGT backbone. Given sampled video frames, this encoder produces 3D-aware features that retain scene geometry. These features are then linearly projected to match both the dimensionality and the effective batch size of features from a conventional 2D visual encoder. The two streams are concatenated and passed through a modality bridge—a lightweight MLP—that converts them into unified visual tokens, which are consumed alongside text tokens by a shared LLM backbone. This geometry-preserving, spatio-temporal pathway yields consistent gains on spatial benchmarks, reporting \textbf{35-45\%} relative improvements over strong baselines.
Similarly, \textsc{Spatial-ORMLLM}~\cite{he2025spatialormllm} incorporates a Spatial-Enhanced Feature Fusion block within the vision tower to inject 3D understanding. In this design, 2D image features are combined with rich 3D cues (\eg, depth or volumetric estimates obtained via an external algorithm) inside a fusion module, and the resulting 2D+3D feature is fed into the LLM’s visual encoder. This end-to-end architecture effectively endows the model with volumetric spatial reasoning using only monocular RGB input, achieving robust 3D scene understanding in complex environments (like surgical operating rooms) without additional sensors.
Another notable system, \textsc{SpatialRGPT}~\cite{cheng2024spatialrgpt}, integrates spatial reasoning capabilities by adding a plug-in depth module and leveraging region-level training signals. In particular, SpatialRGPT uses a ``flexible'' depth-integration module that attaches to the existing visual encoder, enabling it to process inferred depth maps alongside RGB features. Moreover, it is trained with a curated pipeline of 3D scene-graph data to learn detailed regional representations, which allows the model to interpret user-provided region proposals and accurately judge their relative directions and distances during inference. This yields marked improvements in spatial question-answering, both with and without explicit region prompts. Yet another architectural innovation is found in \textsc{Cambrian-1}~\cite{tong2024cambrian}, a vision-centric multimodal model that introduces a Spatial Vision Aggregator (SVA). The SVA is a dynamic, spatially-aware connector module that fuses high-resolution visual feature maps into the LLM while intelligently reducing the number of visual tokens required. By preserving fine-grained spatial information from the vision encoder and feeding it more efficiently to the language model, Cambrian-1 achieves better visual grounding and overall multimodal performance (it served as an open-source testbed that reached state-of-the-art results on a new CV-Bench benchmark). 

Across these designs, the common theme is the addition of structural bias for space: by introducing new layers or networks devoted to geometric processing (be it via explicit spatial feature fusion, graph relationships, or high-res feature aggregation), the models can maintain spatial layouts through the reasoning process, instead of relying solely on implicit signals in the image embeddings.

\noindent \textit{\textbf{Insights \& Discussion.}}  
Dedicated modules inject geometric inductive bias: multi-scale encoders, relation graphs, and spatial cross-attention preserve layout/topology; domain-tailored 2D+3D fusion and depth-integrated connectors enhance robustness under occlusion and clutter. Furthermore, vision-centric aggregators retain fine spatial detail with fewer tokens, and aligning static 3D context with video stabilizes temporal grounding. Nevertheless, added complexity, latency, and reliance on pseudo-3D labels motivate intent-aware routing between spatial modules and the LLM, unified 2D/3D/temporal consistency objectives, and lightweight hardware-friendly spatial layers for deployment.

\subsection{Explainability of Multimodal Spatial Reasoning}
Understanding why MLLMs struggle with spatial reasoning is essential for advancing their design and interpretability. Recent studies have provided valuable insights into these limitations and suggested strategies for improvement.  

From a mechanistic perspective, Rajabi~\etal~\cite{rajabi2024towards} reveal through attention visualization that current MLLMs often rely on object co-occurrence rather than genuine geometric grounding. To address this, they propose decomposing spatial descriptions into grounded subject–object–relation triplets, linking detection and positional features through a lightweight relational bridge. 

Following this thread, Qi~\etal~\cite{qi2025beyond} identify a representational imbalance in multimodal Transformers where dominant vision embeddings suppress positional encodings, erasing spatial order. Using interpretability metrics, they attribute this to cross-modal norm disparities and propose normalizing vision token magnitudes and injecting mid-layer geometric features to recover spatial sensitivity without altering the backbone.

Chen~\etal~\cite{chen2025why} further analyze attention maps and found that only 15–20\% of attention weights target regions encoding spatial relationships, indicating that MLLMs focus on isolated objects instead of inter-object relations. They propose \textit{ADAPTVIS}, a training-free inference strategy that dynamically adjusts attention based on confidence, helping the model refocus on relevant spatial regions. This process-level modulation highlights attention control as an effective route to better spatial grounding.

In parallel, Wen~\etal~\cite{wen2024transformers} show that even large MLLMs often depend on bounding-box heuristics instead of genuine relational cues. They recast spatial relation prediction as a global object–object interaction problem and introduce \textit{RelatiViT}, a transformer that integrates relation-awareness directly into self-attention, embedding structural bias for spatial reasoning into the encoder itself.  

Finally, Zhang~\etal~\cite{zhang2025struggle} take a broader view, showing that simply scaling multimodal data yields diminishing gains on spatial reasoning tasks. Their analysis indicates that spatial competence relies more on the positional fidelity of the vision encoder than on the LLM’s textual positional signals. They advocate embedding explicit 3D-aware modules and cross-view fusion layers to ensure spatial understanding emerges from structure rather than scale.  

\noindent \textit{\textbf{Insights \& Discussion.}}  
Together, these studies converge on a shared diagnosis: MLLMs exhibit strong semantic reasoning but weak spatial grounding due to representational imbalance, attention bias, and lack of geometric priors, which emphasize the need for models that balance semantic and spatial representations. 
Future research should focus on integrating these complementary insights—explicit spatial grounding, balanced cross-modal encoding, relation-aware attention, and geometry-informed architectural priors—to enhance the accuracy and robustness of MLLMs in reasoning about spatial configurations.

    \begin{table*}[h!]
    \centering
    \resizebox{\textwidth}{!}{
        \begin{tabular}{l|l|l|l|p{9cm}} 
            \toprule
            Year & Method & Input & Backbone & Highlights \\ 
            \midrule
            2023 & LLM-Grounder~\cite{yang2024llm} & Point Cloud & GPT-4 & Uses LLM as an agent for 3D closed-loop, feedback-driven visual grounding, which is fully zero-shot and open-vocabulary \\ \midrule
            2023 & Grounded 3D-LLM~\cite{chen2024grounded} & Point Cloud & Tiny-Vicuna-1B & Unifies 3D task modeling with LLM \\ \midrule
            2024 & Vigor~\cite{wu2025data} & Point Cloud & GPT-3.5-Turbo & Introduces referential order modeling for language-obj structure\\ \midrule
            2023 & ViewRefer~\cite{guo2023viewrefer} & Multi-View Image& GPT-3 & Multi-view modeling improves spatial perception \\ \midrule
            2023 & 3DAxiesPrompts~\cite{liu20253daxisprompt} & Multi-View Image& GPT-4V & First to encode 3D coordinates into prompt input \\ \midrule
            2024 & VLM-Grounder~\cite{xu2024vlm} & Multi-View Image& GPT-4V & Utilizes dynamic stitching strategy that dynamically uses
            the optimal layouts to stitch images, enhancing VLM’s performance \\ \midrule
            2024 & SpatialRGPT~\cite{cheng2024spatialrgpt} & RGB-D& LLaMA2-7B & Modular design enables flexible integration \\ \midrule
            2024 & ZSVG3D~\cite{yuan2024visual} & RGB-D & GPT3.5 & First use of program generation in 3DVG \\ \midrule
            2024 & SeeGround~\cite{li2024seeground} & Text+RGB+3D & Qwen2-VL-72B & Dynamically adjusts perspectives to capture essential details \\ \midrule
            2025 & ReasonGrounder~\cite{liu2025reasongrounder} & RGB+3DGS& LLaVA 1.5 & Integrates LVLM, 3DGS, and hierarchical features, enables amodal perception under occlusion\\
            \bottomrule
        \end{tabular}
    }
    \caption{Comparison of recent multimodal spatial reasoning methods in 3D Grounding.}
    \label{tab:3D_grounding}
\end{table*}

\begin{figure*}[h!]
  \centering
  \includegraphics[width=0.95\linewidth]{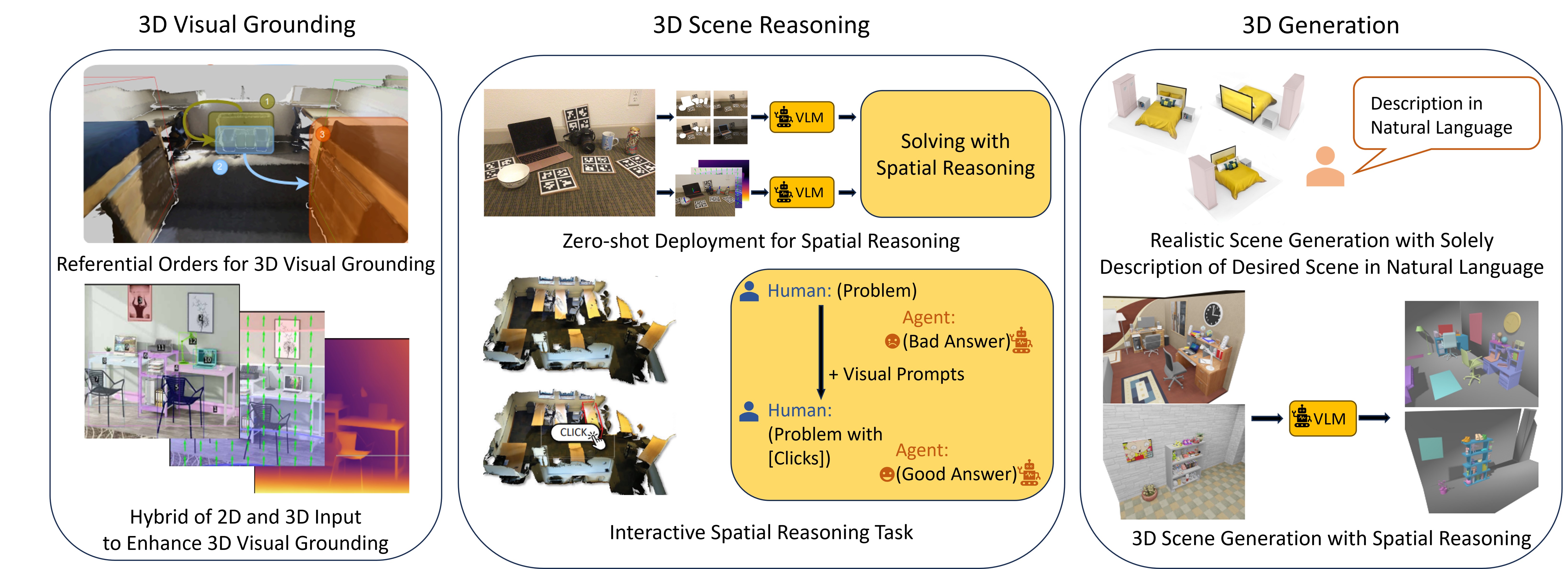}
  \caption{
     An overview of core spatial reasoning tasks in 3D space, including 3D visual grounding\cite{wu2025data, cheng2024spatialrgpt}, 3D scene reasoning\cite{ma2024spatialpin, chen2024ll3da}, and 3D generation\cite{ocal2024sceneteller, wu2024diorama}.
  }
  \label{fig:3D}
\end{figure*}
\begin{figure}[h!]
    \centering
    \includegraphics[width=1\linewidth]{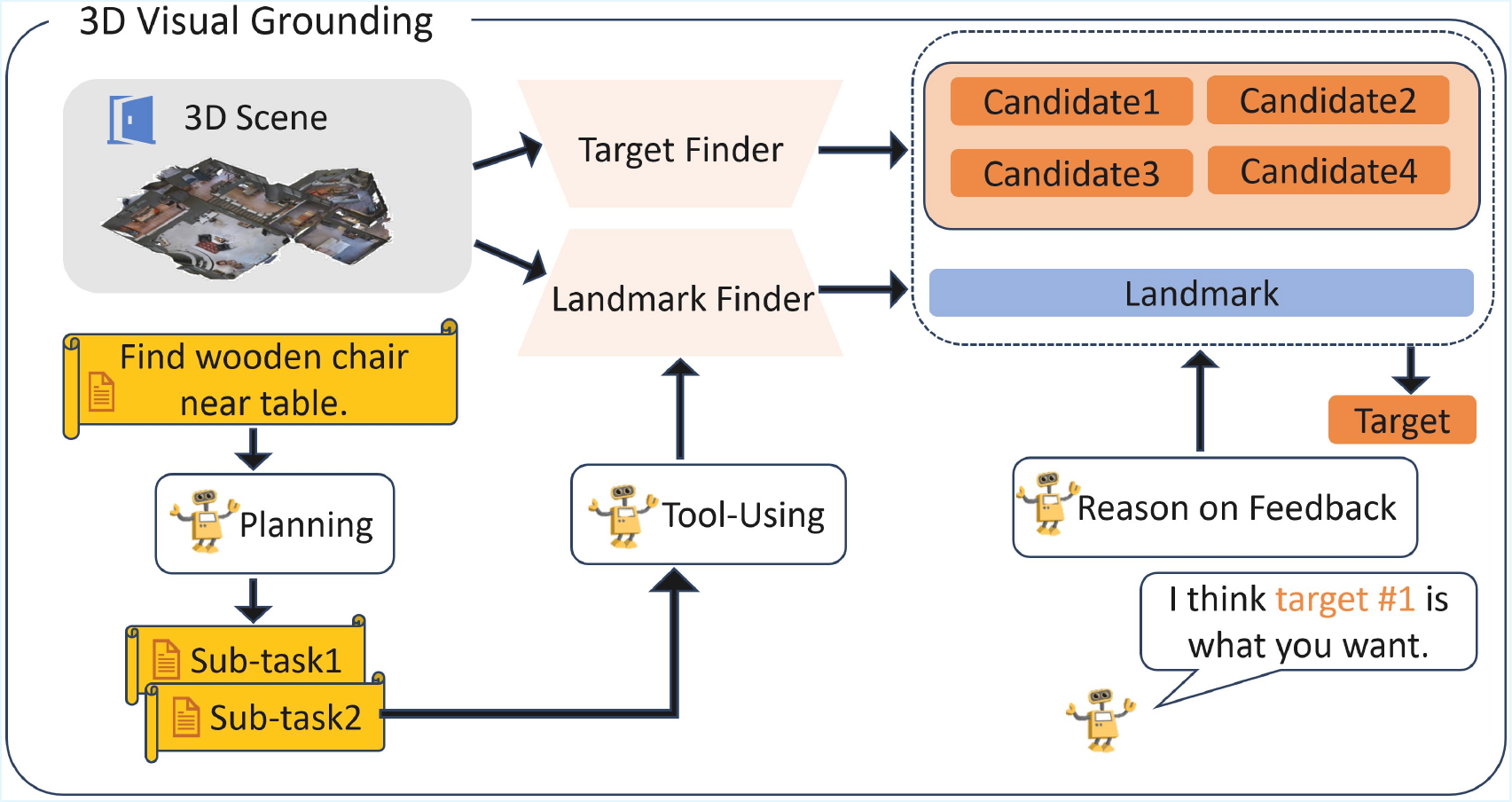}
    \caption{3D visual grounding with MLLM~\cite{yang2024llm}.}
    \label{fig:llmgrounder}
\end{figure}

\section{Multimodal Spatial Reasoning in 3D Space}
Multimodal spatial reasoning in 3D space is a key area of research, with significant implications for downstream applications such as navigation~\cite{yin2024navigation, lin2025navcot}, vision-language-action tasks~\cite{wen2025dexvla, zheng2024tracevla}, and more. This section focuses on foundational tasks with multimodal spatial reasoning, including 3D grounding, 3D scene reasoning, and 3D generation. As illustrated in Figure~\ref{fig:3D}, we provide an overview of these core tasks, highlighting their roles within the broader landscape of 3D spatial understanding.

\subsection{3D Visual Grounding}
As in Figure~\ref{fig:llmgrounder}, given a natural language description, 3D grounding involves localizing an object in a 3D scene. This task requires strong spatial reasoning to handle complex instructions and is crucial for robotics and AR, combining language understanding and 3D spatial reasoning. Traditional 3D grounding methods are fully supervised on limited 3D datasets with predefined object captions~\cite{zhao20213dvg}, but they struggle to generalize to unseen objects and handle complex texts.

Unlike traditional methods, researchers are developing approaches based on MLLMs, significantly enhancing generalizability by leveraging large-scale priors. However, integrating MLLMs into 3D grounding remains challenging~\cite{huang2025unveiling}. Existing approaches for embedding MLLMs into 3D grounding systems can be broadly categorized based on the input data modality:
\ding{172} direct utilization of 3D representations and spatial information;
\ding{173} generation of multi-view 2D images rendered from 3D scenes;
\ding{174} hybrid methods combining both 2D and 3D modalities, as shown in Table~\ref{tab:3D_grounding}.

\subsubsection{3D Input}
Some methods perform spatial reasoning by embedding 3D formats—such as point clouds, voxels, or learned volumetric features—into MLLMs~\cite{yang2024llm, chen2024grounded, wu2025data}.
LLM-Grounder~\cite{yang2024llm} adopts a coarse-to-fine approach, first using an MLLM to parse complex linguistic concepts and an open-vocabulary 3D vision module to generate candidate proposals, then evaluating their semantic alignment with the query. Grounded 3D-LLM~\cite{chen2024grounded} integrates scene-referent tokens into the MLLM and employs alignment training to enable 3D input, leveraging the MLLM's reasoning capabilities. Vigor~\cite{wu2025data} focuses on interpreting spatial language by using an LLM to infer the referential order of entities, enhancing fine-grained spatial reasoning.

\noindent \textit{\textbf{Insights \& Discussion.}} In summary, these approaches focus on 3D visual grounding by embedding 3D representations into MLLMs and utilizing their spatial reasoning ability. However, while embedding 3D modalities holds great potential, it presents challenges. The complexity of 3D data structures can hinder model interpretability, and the limited availability of labeled 3D datasets constrains the development of robust, generalizable models for open-world applications.

\subsubsection{Multi-view Input}

While 3D point clouds provide explicit scene representation, they present challenges for models due to the complexity of spatial information. To address this, researchers are increasingly adopting multi-view 2D representations as a promising alternative. This approach leverages the spatial reasoning capabilities of existing 2D MLLMs with minimal modifications. Representative methods include ViewRefer~\cite{guo2023viewrefer}, VLM-Grounder~\cite{xu2024vlm}, and 3DAxisPrompt~\cite{liu20253daxisprompt}.

A key challenge in multi-view 3D visual grounding is view discrepancy, which arises from the misalignment between the model's perspective and the source of the grounding instruction. Several methods have been proposed to mitigate this issue. For example, ViewRefer~\cite{guo2023viewrefer} introduces learnable multi-view prototypes to capture inter-view relationships and enable knowledge transfer. VLM-Grounder~\cite{xu2024vlm} dynamically stitches image sequences and incorporates a grounding-and-feedback mechanism. 3DAxisPrompt~\cite{liu20253daxisprompt} enhances the real-world scene by inserting 3D coordinate axes.

\noindent \textit{\textbf{Insights \& Discussion.}} These works leverage powerful MLLMs to align with 3D scenes using 2D multi-view inputs. However, key challenges remain~\cite{cheng2024spatialrgpt}: First, MLLMs designed for global image understanding struggle with parsing specific object regions. Second, spatial perception extends beyond RGB data and requires geometric information like depth or spatial coordinates.

\subsubsection{Hybrid of 2D and 3D}

To combine the advantages of both 3D and multi-view representations, recent methods utilize hybrid inputs, including~\cite{zhu2024scanreason, li2024seeground, cheng2024spatialrgpt, yuan2024visual}.  
SpatialRGPT~\cite{cheng2024spatialrgpt} highlights the limitations of MLLMs relying solely on RGB pixels for 3D tasks. It proposes integrating relative depth maps from depth prediction models with RGB images to enhance spatial perception and reasoning.  
ZSVG3D~\cite{yuan2024visual} defines a visual program interface to standardize spatial relationships, enabling reasoning plans for grounding.  
SeeGround~\cite{li2024seeground} integrates 2D visuals with explicit 3D spatial descriptions to improve object localization.  
3D-MOOD~\cite{yang20253dmood} achieves monocular open-set 3D object detection via lifting the open-set 2D detection into 3D space.
ReasonGrounder~\cite{liu2025reasongrounder} introduces 3D Gaussian splatting features as intermediate representations from SAM~\cite{kirillov2023segment} and CLIP~\cite{radford2021learning}.

\noindent \textit{\textbf{Insights \& Discussion.}} These methods demonstrate the limitations of using only 2D or 3D representations and propose strategies for integrating both modalities. Combining multi-view images and 3D structures enhances performance and robustness in 3D visual grounding systems.

\subsection{3D Scene Reasoning and Question Answering (QA)}

3D scene reasoning and QA require models capable of processing 3D representations—such as point clouds, meshes, neural radiance fields, or multi-view RGB-D inputs—and generating natural language responses grounded in the spatial and semantic structure of the environment. Current research falls into two paradigms: training-required and training-free. Training-required methods fine-tune MLLMs, typically via Q-Former~\cite{chen2024ll3da,qi2024gpt4point} or projection-layer modules~\cite{huang2023embodied,fu2024scene}. Training-free methods use frozen MLLMs with progressive prompting~\cite{ma2024spatialpin} and chain-of-thought reasoning~\cite{ma2024spatialpin,zantout2025sort3d}.

\begin{table}[t!]
    \centering
    \tabcolsep=0.6cm
    \resizebox{\columnwidth}{!}{
        \begin{tabular}{l|l|l} 
            \toprule
            Year & Method & Alignment Technique \\ 
            \midrule
            2023 & Chat-3D~\cite{wang2023chat} & Multi-modal Transformer  \\ \midrule
            2023 & Chat-Scene~\cite{huang2023chat} & Multi-modal Transformer \\ \midrule
            2023 & 3D-LLM ~\cite{hong20233d}& Q-Former-liked module \\
            \midrule
            2023 & GPT4Point~\cite{qi2024gpt4point} & Q-Former-liked module \\
            \midrule
            2024 & LL3DA~\cite{chen2024ll3da} & Q-Former-liked module \\
            \midrule
            2023 & LEO~\cite{huang2023embodied} & LLaVA-liked module \\
            \midrule
            2024 & Scene-LLM~\cite{fu2024scene} & LLaVA-liked module \\
            \midrule
            2024 & LLaVA-3D~\cite{zhu2024llava} & LLaVA-liked module \\
            \midrule
            2025 & 3D-LLaVA~\cite{deng20253d} & LLaVA-liked module \\
            \bottomrule
        \end{tabular}
    }
    \caption{Comparison in alingment methods.}
    \label{tab:comparion_of_alignment_techniques}
\end{table}
\subsubsection{Training-required}

Training-required studies can be classified into three categories:
\ding{172} \textbf{Alignment approach}: These methods focus on aligning 3D features with language modalities.
\ding{173} \textbf{Training efficiency}: Aiming to reduce complexity and improve convergence.
\ding{174} \textbf{3D Representation}: Expanding beyond conventional 3D representations to scene graphs, 3DGS~\cite{kerbl20233d,ma2025large}, etc.

The next sections elaborate on each category, summarizing current advancements in multimodal spatial reasoning for 3D.

\ding{172} Recent methods focus on aligning 3D scene features with MLLM feature spaces. Early works~\cite{wang2023chat, huang2023chat} use 3D detectors to extract object-level representations, which are aligned with text features using 3D-text paired data, enabling MLLMs to leverage prior knowledge. However, reliance on 3D detectors can be a bottleneck. To address this, inspired by Q-Former~\cite{li2023blip}, recent works~\cite{chen2024ll3da, hong20233d, qi2024gpt4point, xiong20253ur} integrate similar designs into 3D MLLMs for more complex reasoning. For example, 3UR-LLM~\cite{xiong20253ur} uses a 3D compressor to condense 3D features into compact vision tokens and a 3D query fusion mechanism to select high-confidence queries, improving reasoning robustness.

Besides Q-Former, several methods~\cite{huang2023embodied, fu2024scene, zhu2024llava, deng20253d} are inspired by LLaVA. These approaches use a projection layer to align the feature space with LLMs, enabling them to process 3D inputs and leverage their spatial reasoning capabilities. For example, Scene-LLM~\cite{fu2024scene} employs a two-stage strategy, training a projection layer with conceptual annotations while keeping the LLM frozen. An overview of these alignment techniques is presented in Table~\ref{tab:comparion_of_alignment_techniques}.

\ding{173} Beyond improving alignment quality, recent studies~\cite{li20243dmit, zhu2024llava, deng20253d, yu2025inst3d} note that aligning 3D features with language is time-consuming. To improve efficiency, 3DMIT~\cite{li20243dmit} removes the alignment step by focusing on instruction tuning for spatial understanding. LLaVA-3D~\cite{zhu2024llava} retains LLaVA's 2D multimodal capabilities by constructing 3D patches and using 3D-aware positional encoding. Inst3D-LMM~\cite{yu2025inst3d} introduces multi-task instruction tuning, enabling adaptation to various spatial reasoning tasks without task-specific fine-tuning.

\begin{table}[t!]
    \centering
    \tabcolsep=0.3cm
    \resizebox{\columnwidth}{!}{
        \begin{tabular}{c|l|c|c} 
            \toprule
            Year & Method & Representation & Training \\ 
            \midrule
            2024 & 3DGraphLLM~\cite{zemskova20243dgraphllm} & Scene Graph & Full Training \\ \midrule
            2025 & SplatTalk~\cite{thai2025splattalk} & 3DGS & Fine-tuning \\ \midrule
            2025 & GPT4Scene~\cite{qi2025gpt4scene} & BEV & Zero-shot / Fine-tuning \\
            \bottomrule
        \end{tabular}
    }
    \caption{Comparison of multimodal spatial reasoning methods with diverse 3D representations.}
    \label{tab:Comparison_of_3D_representations}
\end{table}

\ding{174} Recent works~\cite{zemskova20243dgraphllm, thai2025splattalk, qi2025gpt4scene} focus on diverse 3D representations, including 3D scene graphs, 3DGS, and BEV. 
3DGraphLLM~\cite{zemskova20243dgraphllm} creates a learnable 3D scene graph to enhance spatial reasoning by utilizing richer structural information. 
SplatTalk~\cite{thai2025splattalk} integrates language features from RGB images into a unified 3DGS~\cite{kerbl20233d} representation, supporting spatial reasoning. 
GPT4Scene~\cite{qi2025gpt4scene} improves reasoning by reconstructing BEV images from 3D scene videos and establishing a consistent mapping between local views and global scene structure. 
A comparison of these 3D representations is provided in Table~\ref{tab:Comparison_of_3D_representations}.

\noindent \textit{\textbf{Insights \& Discussion.}} Efforts to enhance 3D spatial reasoning in MLLMs focus on modality alignment, training efficiency, and exploring alternative 3D representations. However, challenges remain: 
\ding{172} Training 3D-aware models is computationally intensive due to complex data and architectures.
\ding{173} The lack of large, diverse, and well-annotated 3D datasets limits the effectiveness of supervised training.
\ding{174} The absence of transparent reasoning mechanisms hinders interpretability and understanding of model decisions.
Addressing these limitations could further advance MLLMs for spatial reasoning.

\begin{figure*}[h!]
  \centering
  \includegraphics[width=0.95\linewidth, trim=0 0 0pt 0, clip]{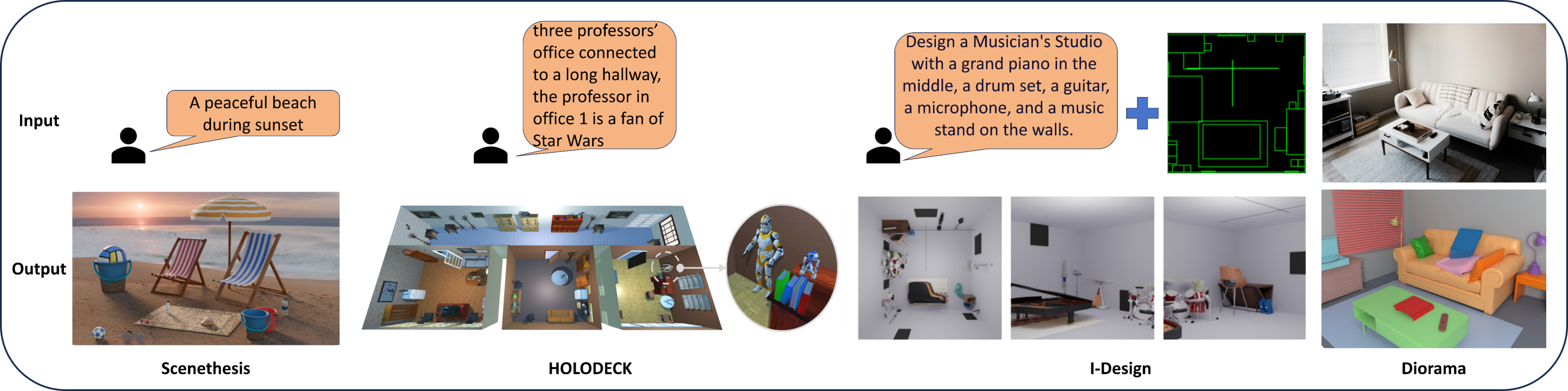}
  \caption{
     Some comparative examples of 3D generations, such as input conditions (e.g., text or image), and outputs from different approaches\cite{ling2025scenethesis, yang2024holodeck, ccelen2025design, wu2024diorama}, showcasing variations in geometry, texture, and semantic coherence.
  }
  \label{fig:3d-generation-comparison}
\end{figure*}

\subsubsection{Training-free Methods}

Training-free methods~\cite{ma2024spatialpin,zhang2024agent3d,he2024think,zantout2025sort3d} leverage the prior knowledge in MLLMs for multimodal spatial reasoning without the need for fine-tuning. These methods explore various prompting strategies to facilitate interpretable spatial reasoning. Some works~\cite{ma2024spatialpin,zantout2025sort3d} use MLLMs to extract semantic object attributes and apply the chain-of-thought mechanism, prompting sequential reasoning. SpatialPIN~\cite{ma2024spatialpin} is a modular framework that employs progressive prompting to decompose and reconstruct explicit 3D representations, enhancing spatial reasoning. Agent3D-Zero~\cite{zhang2024agent3d} introduces a Set-of-Line strategy for selecting and analyzing multiple viewpoints, improving spatial reasoning while reducing memory and computation. LLM-TPC~\cite{he2024think} employs a Think-Program-reCtify loop to bridge 3D visual perception and reasoning, improving reliability through iterative self-correction.

\noindent \textit{\textbf{Insights \& Discussion.}} These training-free methods utilize MLLMs to summarize and refine spatial information through diverse prompting strategies. Despite their success, they have limitations: \ding{172} They depend on the quality of the MLLMs used, and deficiencies in these models may hinder performance on some tasks. \ding{173} Some methods involve complex inference steps, reducing processing speed and making them less suitable for real-time applications.

\subsection{3D Generation with Spatial Reasoning}

3D generation~\cite{jiang2023sdf, jiang2025dimer} has advanced rapidly, particularly with the integration of LLMs and multimodal reasoning systems. Scene-level and program-level generation demand strong spatial reasoning capabilities. These tasks can be categorized into two aspects:
\ding{172} 3D Layout Generation: Generating spatially reasonable indoor layouts from natural language or multi-turn dialogues.
\ding{173} 3D Generation as Program: Treating 3D content generation as a programmatic task, where spatial reasoning is framed as executable program generation.

\subsubsection{3D Layout Generation}
Given the complexity of 3D scene generation~\cite{jiang2024general, jiang2024brightdreamer, hua2025sat2city}, researchers often use MLLMs for initial 3D layout generation, followed by scene-level synthesis. 
Figure~\ref{fig:3d-generation-comparison} presents a qualitative comparison of representative 3D scene generation approaches, showcasing variations in geometric fidelity, texture quality, and semantic consistency across different methods. Approaches can be broadly categorized based on how MLLMs are integrated into the layout pipeline:

\ding{172} \textbf{Direct Guidance for Scene Synthesis via LLMs}: MLLMs directly generate spatial configurations or layout instructions, translating high-level descriptions into structured commands for scene elements, such as furniture arrangement and room dimensions. However, this direct mapping can lead to implausible configurations, like overlapping objects. Methods like LayoutGPT~\cite{feng2023layoutgpt} and HOLODECK~\cite{yang2024holodeck} address this by incorporating optimization-based solvers or inferring spatial relational constraints.

\ding{173} \textbf{Indirect Guidance for Scene Synthesis via LLMs}: 
Indirect guidance uses MLLMs to extract semantic knowledge (e.g., object relationships or contextual constraints) to guide subsequent 3D modeling. For instance, Diorama~\cite{wu2024diorama} generates a scene graph defining object relationships, while the MLLM retrieves multimodal 3D shapes. Approaches like LayoutGPT~\cite{feng2023layoutgpt} use programmatic reasoning to generate spatial layout specifications, while HOLODECK~\cite{yang2024holodeck} enhances this with optimization techniques for physical realism. Iterative methods, such as I-Design~\cite{ccelen2025design} and Generation Agents~\cite{sasazawa2024layout}, introduce multi-agent systems for step-by-step refinement. LLPlace~\cite{yang2024llplace} supports real-time interactive layout refinement through a conversational interface, and Chat2Layout~\cite{wang2024chat2layout} combines VQA with visual prompting, enhancing spatial layout reasoning.

\noindent \textit{\textbf{Insights \& Discussions.}} The primary approaches either generate positions directly or create intermediate representations like scene graphs. Both paradigms leverage MLLMs for semantically coherent and physically feasible 3D environments. Future advancements in MLLMs could enhance both numerical accuracy and formatting capabilities.

\begin{figure}
    \centering
    \includegraphics[width=1\linewidth]{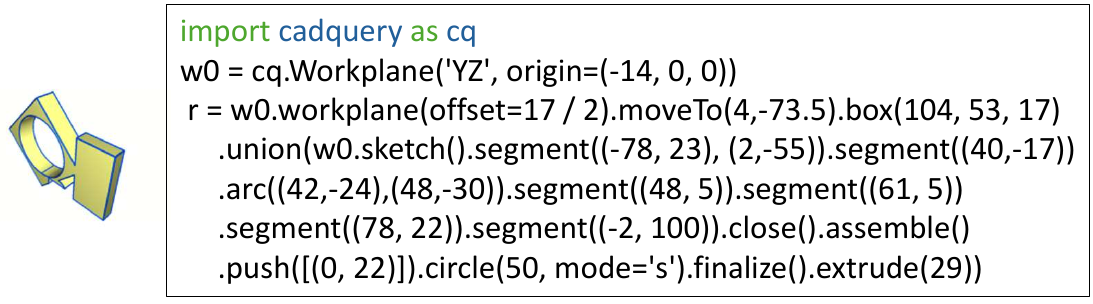}
    \caption{A demo for programming 3D object representation. The left is a CAD model, and the right is the corresponding code segment of CAD Query~\cite{rukhovich2024cad}.}
    \label{fig: 3DGen_as_program_demo}
\end{figure}

\subsubsection{3D Generation as Program}

Building on advances in MLLM-based code generation (e.g., Cursor~\cite{cursor} and GitHub Copilot~\cite{githubcopilot}), recent work treats 3D synthesis as procedural \emph{program} generation, where geometry and layout are specified by code. As shown in Fig.~\ref{fig: 3DGen_as_program_demo}, a 3D model can be described by a code snippet, leveraging MLLMs’ structured reasoning and constraints. Current approaches target three output formats: \ding{172} Blender scripts, \ding{173} CAD parametric programs, and \ding{174} mesh-generation pipelines.

\ding{172} Blender is the most common software in 3D modeling and animation, supporting operations via its API and Python code. The following methods utilize MLLMs' spatial reasoning for programming outputs. 3D-GPT~\cite{sun20233d} introduces a training-free framework where an LLM interprets natural language commands and generates Blender scripts to construct 3D scenes, unlocking the potential of MLLMs in spatial programming. SceneCraft~\cite{kumaran2023scenecraft} proposes a dual-loop optimization system: an inner loop refines scenes using MLLM feedback, while the outer loop accumulates spatial knowledge across iterations, enabling self-evolving capabilities. SceneMotifCoder~\cite{tam2024scenemotifcoder} introduces ``visual programs’’ — structured code representations extracted from example-based demonstrations.

\ding{173} In addition to Blender, other works extend spatial reasoning into CAD modeling. CAD-GPT~\cite{cad-gpt} enhances spatial reasoning by integrating spatial tokens and positional embeddings, enabling accurate generation of CAD sequences from images or text. CAD2PROGRAM~\cite{wang20252d} converts 2D engineering drawings into executable Python scripts using MLLMs. CAD-Recode~\cite{rukhovich2024cad} maps point cloud data into CadQuery scripts via a lightweight encoder and pre-trained MLLM backbone. CAD-LLaMA~\cite{li2025cad} designs a parametric language to better utilize MLLMs’ spatial knowledge.

\ding{174} Other work focuses on general mesh generation using a programmatic approach. ShapeLib~\cite{jones2025shapelib} guides LLMs in constructing libraries through a hybrid human-AI workflow.

\noindent \textit{\textbf{Insights \& Discussions.}}
These works reflect the expanding scope of MLLMs in tackling complex, real-world tasks that require deep spatial reasoning, precise geometric control, and integration with downstream tools. While directly generating 3D representations is challenging, using MLLMs for 3D content generation via programming harnesses their full spatial reasoning potential. Programmatic generation is also more controllable, making it better suited for real applications.

\begin{figure*}[h!]
  \centering
  \includegraphics[width=0.95\linewidth, trim=0 0 0pt 0, clip]{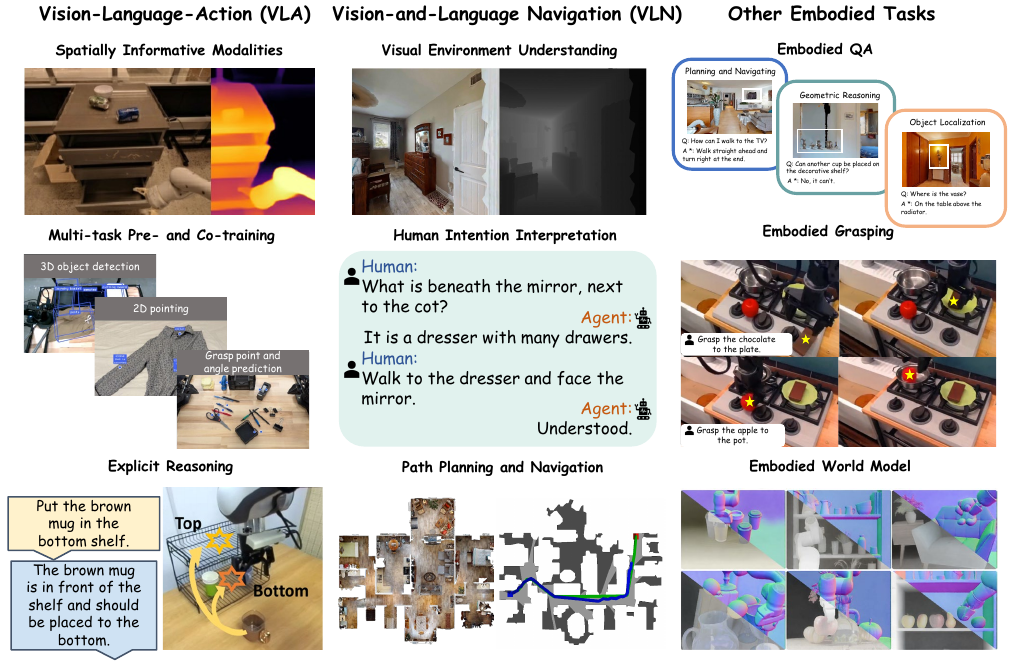}
  \caption{
     Spatial reasoning in embodied tasks, such as VLA~\cite{chat-vla,3d-vla,gemini-robotics}, VLN~\cite{ramakrishnan2021habitat,walke2023bridgedata} and other embodied tasks~\cite{zhen2025tesseract}.
  }
  \label{fig:embodied}
\end{figure*}
\section{Multimodal Spatial Reasoning in Embodied AI}

Embodied AI is regarded as a crucial path toward AGI~\cite{zheng2025panorama}. The rapid progress of MLLMs positions them as promising candidates for the core reasoning module of embodied agents. Many of the core intelligences expected of embodied agents—such as geometric reasoning, navigation, and perspective-taking—fundamentally rely on spatial reasoning capabilities as their foundation~\cite{gardner1989educational,kamath2023s,li2025clivis}.
As demonstrated in Fig.~\ref{fig:embodied}, in this section, we focus on the multimodal spatial reasoning capabilities of MLLM-based embodied agents within the context of current mainstream tasks, including Vision-Language Action (VLA), Vision-and-Language Navigation (VLN), and other embodied AI tasks.

\subsection{Multimodal Spatial Reasoning in VLA Models}
VLA models generate executable actions from multimodal inputs—typically visual observations and language instructions—using vision-language foundation models as their backbone. These systems often involve intermediate reasoning steps, either implicit within the architecture or explicit through modular design. Pioneering works such as OpenVLA~\cite{kimopenvla} and $\pi0$~\cite{pi0} adopt an end-to-end paradigm, training VLMs as reactive policies to predict low-level control actions from large-scale demonstrations. Others~\cite{hirobot, pi05} decompose tasks into natural language sub-tasks executed by reactive controllers or lower-level VLAs, while some frameworks introduce intermediate stages like affordance or goal-state prediction followed by motion planning for action generation.  

Regardless of the control representation, spatial reasoning remains central to these systems. Research efforts to improve spatial understanding in VLAs generally follow three directions: \ding{172} integrating spatially informative sensor modalities (e.g., depth, point clouds) to enrich spatial context; \ding{173} adopting multi-task pre-training or co-training schemes that implicitly encourage spatial reasoning; and \ding{174} incorporating explicit reasoning steps. The following subsections review representative methods in each direction and discuss their respective advantages and limitations.

\subsubsection{Spatially informative input modalities}
\begin{table}[t]
\centering
\caption{Comparison of 3D-enhanced VLA methods. \checkmark indicates the feature is present, and \xmark indicates it is absent.}
\label{tab:3dvla_comparison}
\setlength{\tabcolsep}{3pt}
\resizebox{\linewidth}{!}{
\begin{tabular}{l|cccccc}
\toprule
Method &
\rotatebox{90}{3D Perception} &
\rotatebox{90}{Depth Maps} &
\rotatebox{90}{Point Clouds} &
\rotatebox{90}{Goal Generation} &
\rotatebox{90}{Spatial Encoding} &
Training Strategy \\
\midrule
\textbf{3D-VLA}~\cite{3d-vla} & \checkmark & \checkmark & \checkmark & \checkmark & \xmark & Diffusion-aligned \\
\textbf{PointVLA}~\cite{pointvla} & \checkmark & \xmark & \checkmark & \xmark & \xmark & Action expert fusion \\
\textbf{SpatialVLA}~\cite{qu2025spatialvla} & \checkmark & \checkmark & \xmark & \xmark & \checkmark & Monocular depth-based encoding \\
\textbf{BridgeVLA}~\cite{bridge-vla} & \checkmark & \xmark & \checkmark & \xmark & \checkmark & Dual-phase (2D pre-train + 3D fine-tune) \\
\bottomrule
\end{tabular}}
\end{table}
Several studies enhance spatial understanding in VLA models by incorporating spatially informative modalities such as depth maps and 3D point clouds, as shown in Table~\ref{tab:3dvla_comparison}. These additional inputs compensate for the limitations of 2D visual data, which often lack the geometric cues needed for reasoning about physical interactions in 3D space.
3D-VLA~\cite{3d-vla} enhances a language model with 3D perception and goal generation by introducing interaction tokens for objects, locations, scenes, and actions. It aligns the language model with diffusion models that generate goal images, depth maps, and point clouds from instructions. PointVLA~\cite{pointvla} combines 2D image features from a VLM and 3D point cloud features from a point encoder as inputs to an action expert for prediction. SpatialVLA~\cite{qu2025spatialvla} encodes 3D information into 2D observations using 3D-aware positional encodings derived from monocular depth predictions. BridgeVLA~\cite{bridge-vla} employs dual-phase training: pre-training a VLM for 2D heatmap-based object localization and fine-tuning with multi-view orthographic projections of 3D point clouds to generate action trajectories.


\noindent \textit{\textbf{Insights \& Discussion.}} These approaches show promise for action prediction with richer spatial perception, but challenges remain. A key limitation is the scarcity of large-scale datasets compared to vision–language corpora~\cite{pointvla,bridge-vla}, motivating synthetic data~\cite{3d-vla} or imputing missing modalities with pre-trained models (e.g., SpatialVLA~\cite{qu2025spatialvla}). Yet such approximations often underperform. Moreover, models trained at scale on 2D vision–language data still lead overall~\cite{gemini-robotics,pi0,pi05}, indicating that fully leveraging extra modalities will require targeted pre-training and more data-efficient architectures.

\begin{table}[t]
\centering
\caption{Comparison of multi-task pre- and co-training strategies for VLA models. \checkmark indicates the feature is present, and \xmark indicates it is absent.}
\label{tab:multitask_vla_comparison}
\renewcommand{\arraystretch}{1.2}
\setlength{\tabcolsep}{3pt}
\resizebox{\linewidth}{!}{
\begin{tabular}{l|
>{\centering\arraybackslash}m{1.1cm}|
>{\centering\arraybackslash}m{1.1cm}|
>{\centering\arraybackslash}m{1.1cm}|
>{\centering\arraybackslash}m{1.1cm}|
>{\centering\arraybackslash}m{1.1cm}}
\toprule
\textbf{Method} &
\rotatebox{90}{\textbf{Embodied QA}} &
\rotatebox{90}{\textbf{3D Tasks}} &
\rotatebox{90}{\textbf{Trajectory Pred.}} &
\rotatebox{90}{\textbf{Multi-Task Co-train}} &
\rotatebox{90}{\textbf{Curriculum / Stage}} \\
\midrule
\textbf{RT-2}~\cite{RT-2} & \checkmark & \xmark & \xmark & \checkmark & \xmark \\
\textbf{Gemini Robotics}~\cite{gemini-robotics} & \checkmark & \checkmark & \checkmark & \checkmark & \checkmark \\
\textbf{$\pi0.5$}~\cite{pi05} & \checkmark & \xmark & \checkmark & \checkmark & \checkmark \\
\textbf{ChatVLA}~\cite{chat-vla} & \checkmark & \xmark & \xmark & \checkmark & \checkmark \\
\textbf{Magma}~\cite{magma} & \xmark & \xmark & \xmark & \checkmark & \xmark \\
\bottomrule
\end{tabular}}
\end{table}

\subsubsection{Multi-task Pre- and Co-training}
Another major approach to enhance spatial understanding in VLA models is to modify the training regime to include auxiliary tasks that implicitly encourage spatial reasoning, such as embodied question answering or 3D bounding box detection, as in Table~\ref{tab:multitask_vla_comparison}. This is typically achieved through pre-training or co-training frameworks that share representations across related spatial tasks. The concept is first explored in \textsc{RT-2}~\cite{RT-2}, which jointly trained a VLM on visual question answering and robot action prediction within a shared token space. Building on this idea, recent large-scale models like \textsc{Gemini Robotics}~\cite{gemini-robotics} and $\pi0.5$~\cite{pi05} employ multi-stage co-training pipelines. \textsc{Gemini Robotics}~\cite{gemini-robotics} adopts a two-stage procedure: the base VLM is first pre-trained on tasks including trajectory prediction, multi-view correspondence, and 3D bounding box detection, yielding the embodied reasoning model Gemini-ER capable of few-shot control through in-context learning. The model is then fine-tuned with an action decoder that outputs low-level control commands for complex manipulation tasks. Similarly, $\pi0.5$~\cite{pi05} pre-trains its VLM backbone on a mixture of tasks such as visual question answering, object localization, sub-task prediction, and discrete action generation. During post-training, an additional action head is introduced for continuous control prediction, followed by fine-tuning for both continuous control and sub-task reasoning. 

ChatVLA \cite{chat-vla} introduces a two-stage curriculum where the model first learns control from robot data, then training examples from other tasks, such as VQA, are gradually introduces to preserve alignment with pre-trained VLM representations. It also adopts a Mixture of Experts architecture with task-specific heads to avoid task interference. Magma \cite{magma} proposes to bridge the gap between vision-language and action data via surrogate tasks that require predicting actionable 2D annotations---Set-of-Mark and Trace-of-Mark. This enables joint training on diverse datasets across digital and physical domains using the same output representation.


\noindent \textit{\textbf{Insights \& Discussion.}} Pre- and co- training on spatial reasoning tasks is an effective way to enhance the generalization capabilities of VLA models. However, this approach doesn't come without its challenges. It requires access to large and diverse datasets, and carefully balancing multiple training objectives. Still, when these challenges are addressed, it remains a core strategy for building capable VLA models.

\subsubsection{Explicit Reasoning}

A third line of research enhances spatial reasoning in VLA models by introducing explicit reasoning steps during action generation. Unlike reactive policies~\cite{kimopenvla,pi0,wen2025dexvla} that directly map inputs to actions, these models incorporate structured intermediate representations and multi-step reasoning to interpret spatial relations and plan sub-tasks before executing actions.

ECoT~\cite{ecot} trains VLA models to generate step-by-step reasoning chains grounded in the scene and robot state prior to action prediction. These chains include high-level plans, sub-tasks, object locations, and low-level motions, improving both generalization and interpretability.  
Chat-VLA2~\cite{chat-vla2} builds on ChatVLA by adding a reasoning-following module that aligns generated actions with the backbone’s internal reasoning, yielding better performance on multi-step spatial tasks.  
Chain-of-Affordance~\cite{chain-of-affordance-vla} introduces an affordance-based reasoning process that decomposes tasks into four stages: identifying target objects, selecting grasp points, locating placement regions, and planning trajectories. These affordances, generated at inference time, guide the policy model’s action selection.  
Similarly, RT-Affordance~\cite{rt-affordance} proposes a hierarchical VLA where action generation is conditioned on affordance plans. An affordance prediction model first generates key poses from images and task descriptions, which then guide a reactive VLA to produce low-level control actions.


\noindent \textit{\textbf{Insights \& Discussion.}} Reasoning-augmented models improve robustness, generalization, and interpretability in spatial tasks by explicitly modeling intermediate steps such as object selection, spatial relations, and action planning. This structured reasoning helps policies handle novel objects, scenes, and instructions more effectively than purely reactive baselines. While early methods introduced substantial inference overhead, newer systems mitigate this through selective reasoning and asynchronous pipelines. These trends suggest that the benefits of explicit reasoning can be retained without prohibitive latency, making such models increasingly practical for real-world deployment.

\subsubsection{Multimodal Spatial Reasoning in Vision Language backbone}
Many current VLA models are fine-tuned from VLMs or use them as backbones. These VLAs are claimed to effectively inherit the prior knowledge of these pre-trained models. To quantitatively assess the potential of the upstream VLMs for robotics tasks, we collected open-source VLMs that have been used in VLAs and evaluated them on spatial reasoning benchmarks relevant to embodied scenarios. Specifically, OpenVLA~\cite{kimopenvla} is fine-tuned from Prismatic~\cite{karamchetiPrismaticVLMsInvestigating}, $\pi_0$ is fine-tuned from PaliGemma~\cite{beyer2024paligemma}, TraceVLA~\cite{zheng2024tracevla} is fine-tuned from Phi-3-Vision~\cite{abdin2024phi}, and DexVLA~\cite{wen2025dexvla} uses Qwen-2-VL~\cite{wang2024qwen2} as its backbone. 
As for the benchmarks, Embodied Reasoning QA (ERQA)~\cite{gemini-robotics} is a benchmark specifically designed for evaluating VLMs in embodied environments. It tests the VLM's ability to handle embodied tasks. On the other hand, SpatialEval~\cite{wang2024spatial} and SPACE~\cite{SPACE} are benchmarks that assess the more fundamental and conventional spatial reasoning abilities of VLMs, such as the ability to judge relative spatial positions and distances. Both of these capabilities are crucial for robotics. Therefore, we conducted experiments by testing several VL backbones used in VLA on these benchmarks. As shown in the Tab.~\ref{tab:embodied_results}, it is evident that these backbones exhibit certain spatial reasoning abilities. This is also why these models can achieve strong performance in downstream applications after fine-tuning on robotic datasets.

\begin{table}[t!]
    \centering
    \resizebox{0.49\textwidth}{!}{
    \begin{tabular}{l|cccc}
        \toprule
        \textbf{Benchmark} & \textbf{Prismatic} & \textbf{PaliGemma} & \textbf{Qwen-2-VL} & \textbf{Phi3-Vision} \\
        \midrule
        ERQA~\cite{gemini-robotics} & 32.25 & 27.25 & 32.50 & 34.00 \\
        SpatialEval~\cite{wang2024spatial} &32.13  &29.86  &26.80  &46.46  \\
        SPACE~\cite{SPACE} &23.75  &17.00   &18.75  &26.25       \\
        \bottomrule
    \end{tabular}
    }
    \caption{Embodied-AI-related benchmark results across different VLMs. Note that SpatialEval is tested using VTQA mode(with Vision-Text input).}
    \label{tab:embodied_results}
\end{table}

\subsection{Multimodal Spatial Reasoning in VLN Models}
VLN~\cite{zhang2024vision} is a cooperative multimodal task where an agent navigates 3D environments by following human instructions and communicating in context under ambiguity. It involves four key components: visual perception, language understanding, decision-making, and navigation execution—all requiring strong spatial reasoning. During perception, the agent must localize itself, interpret spatial relationships between objects, and plan an efficient route. Finally, it executes the navigation plan based on these spatial decisions.


\subsubsection{Visual Environment Understanding and Generalization}


\begin{table*}[t!]
    \centering
    \resizebox{\textwidth}{!}{
        \begin{tabular}{l|l|l|l|p{9cm}} 
            \toprule
            Year & Method & Input & Backbone & Highlights \\ 
            \midrule
            2024& ConceptGraphs~\cite{gu2024conceptgraphs}& RGB-D image& LLaVa
& Constructs open-vocabulary 3D scene graphs\\ \midrule
            2024& NaviLLM~\cite{zheng2024towards} & Multi-view RGB image& Vicuna-7B-v0& Uses schema-based instruction to adapt LLMs\\ \midrule
            2025& Spartun3D-LLM~\cite{zhang2024spartun3d}& Point Cloud & GPT4o& Integrates a 3D-based LLM with a spatial alignment module that links 3D objects and relations to text, bridging the 3D-text gap\\ \midrule
            2025& g3D-LF~\cite{wang2024g3d}& RGB-D image& Vicuna-7B-v0& Proposes generalizable 3D-language feature fields\\ \midrule
            2025& SpatialBot~\cite{cai2024spatialbot}& RGB-D image& QWen1.5-0.5B& Introduces depth API to retrieve geometric information\\
            \bottomrule
        \end{tabular}
    }
    \caption{Comparison of recent multimodal spatial reasoning methods in embodied scene understanding.}
    \label{tab:embodied_2}
\end{table*}

For a VLN agent, it is crucial to perceive and interpret its surroundings, anticipate how actions alter the environment, and align perception and decision-making with natural language instructions. This requires understanding spatial arrangements, localizing itself in 3D space, estimating distances between targets and landmarks, retaining spatial information, and tracking environmental changes over time. These abilities collectively depend on strong spatial reasoning, which underpins success in complex vision-and-language navigation tasks.

 
Existing embodied scene perception methods often rely on 3D or 2.5D data to enhance spatial awareness, as summarized in Tab.~\ref{tab:embodied_2}. To better utilize visual inputs, many approaches explicitly preserve spatial features through multiview perception, depth images, or scene graphs. NaviLLM~\cite{zheng2024towards} leverages multiview images to capture all reachable viewpoints from the current position and constructs task-specific schemas for LLM-based action generation. Cai~\etal~\cite{cai2024spatialbot} propose SpatialBot, which uses a depth API to query geometric information from the environment and feed it back into the model, strengthening spatial understanding. ConceptGraphs~\cite{gu2024conceptgraphs} builds an open-vocabulary 3D scene representation by associating 2D foundation model outputs across multiple views.

Beyond visual encoding, another research direction focuses on narrowing the semantic gap between natural language and 3D scene understanding. 
Spartun3D-LLM~\cite{zhang2024spartun3d} integrates a 3D-aware LLM with a situated spatial alignment module to better link 3D visual representations with corresponding textual descriptions. 
Similarly, Wang~\etal~\cite{wang2024g3d} introduce a 3D representation model for embodied tasks that predicts novel views and BEV maps at multiple scales, aligning multi-scale feature fields with multi-granularity language representations.

Beyond scene understanding, maintaining environmental memory and tracking temporal changes are equally important. 
Hong~\etal~\cite{hong2025general} propose GSA-VLN, where agents dynamically update parameters, leverage long-term memory, and adapt to both environments and diverse user instructions. 
Similarly, Yang~\etal~\cite{yang20253d} present 3D-Mem, a memory architecture that encodes multi-view 3D snapshots to accumulate and retrieve spatial information for long-term perception and reasoning.

\begin{figure}[t!]
    \centering
    \includegraphics[width=1\linewidth]{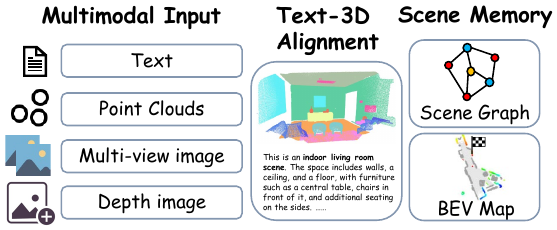}
    \caption{Visual environment understanding in VLN tasks. Current methods take text, point clouds~\cite{zhang2024spartun3d}, multi-view images~\cite{zheng2024towards}, RGB-D images~\cite{gu2024conceptgraphs, cai2024spatialbot, wang2024g3d} as inputs and align them with 3D scene representations, while maintaining structured memories such as scene graphs~\cite{gu2024conceptgraphs} and BEV maps~\cite{wang2024g3d} for effective spatial reasoning. }
    \label{fig:embodied_2}
\end{figure}
\noindent \textit{\textbf{Insights \& Discussion.}} Accurate perception, robust spatial reasoning, and generalization across diverse visual scenes are fundamental for VLN agents. As shown in Fig.~\ref{fig:embodied_2}, recent work emphasizes structured 3D representations, such as scene graphs, BEV maps, and multiview memory, as effective tools linking perception to reasoning and planning. A key challenge remains the alignment of visual features with linguistic inputs, especially under unfamiliar views or domain shifts.

\subsubsection{Human Intention Interpretation and Instruction Comprehension}
VLN agents are required to comprehend natural language instructions provided by humans within specific situational contexts to complete navigation tasks. This involves cor
rectly interpreting spatial expressions such as “left,” “up,” and “front,” and developing the ability to reason spatially about object locations, directions, and movements~\cite{yang2024thinking}. 
To facilitate efficient instruction understanding, a common strategy is to incorporate auxiliary modalities into the input. LL3DA~\cite{chen2024ll3da} encodes 3D point clouds and leverages an attention mechanism to aggregate contextual information from both the scene and human interactions. 

In addition, improved VQA paradigms can further enhance an agent’s instruction comprehension. AutoSpatial~\cite{kong2025autospatial} applies a hierarchical two-round VQA strategy during training, achieving both global and detailed understanding of scenarios, which demonstrates more accurate spatial perception. 

Moreover, certain methods, such as affordance prediction, have been introduced to improve the model’s ability to attend to fine-grained visual details under human instructions.
Yuan~\etal~\cite{yuan2024robopoint} proposed RoboPoint, a vision-language model tailored for predicting spatial affordances from relational language inputs. The model predicts precise action points that comply with spatial and physical constraints, thereby facilitating subsequent action execution.

\noindent \textit{\textbf{Insights \& Discussion.}} Recent work highlights the benefits of auxiliary modalities, hierarchical reasoning, and affordance modeling in improving instruction understanding. Multi-round VQA and affordance prediction enhance fine-grained grounding, while attention-based fusion with human interactions supports contextual comprehension. Future advances may rely on tighter integration of spatial perception and language reasoning, along with better generalization to diverse instructions and complex real-world tasks.

\begin{table}[t]
\centering
\caption{Comparison of path planning and navigation methods for VLN agents. \checkmark indicates feature presence.}
\label{tab:vln_navigation_comparison}
\setlength{\tabcolsep}{1pt}
\resizebox{\linewidth}{!}{
\begin{tabular}{l|
>{\centering\arraybackslash}m{1.3cm}|
>{\centering\arraybackslash}m{1.3cm}|
>{\centering\arraybackslash}m{1.3cm}|
>{\centering\arraybackslash}m{1.3cm}|
>{\centering\arraybackslash}m{1.3cm}|
>{\centering\arraybackslash}m{1.3cm}}
\toprule
\textbf{Method} &
\rotatebox{90}{\textbf{Spatial Reasoning}} &
\rotatebox{90}{\textbf{CoT Reasoning}} &
\rotatebox{90}{\textbf{Domain Adapt.}} &
\rotatebox{90}{\textbf{Hallucination Mitig.}} &
\rotatebox{90}{\textbf{Hierarchical Planning}} &
\rotatebox{90}{\textbf{Mapping/Pre-map}} \\
\midrule
\textbf{NavVLM}~\cite{yin2024navigation} & \checkmark & \xmark & \xmark & \xmark & \xmark & \xmark \\
\textbf{SpatialCoT}~\cite{liu2025spatialcot} & \checkmark & \checkmark & \xmark & \xmark & \xmark & \xmark \\
\textbf{NavCoT}~\cite{lin2025navcot} & \checkmark & \checkmark & \checkmark & \xmark & \xmark & \xmark \\
\textbf{FlexVLN}~\cite{zhang2025flexvln} & \checkmark & \checkmark & \checkmark & \checkmark & \xmark & \xmark \\
\textbf{NavA$^3$}~\cite{zhang2025nava} & \checkmark & \xmark & \xmark & \xmark & \checkmark & \xmark \\
\textbf{TopV-Nav}~\cite{zhong2024topv} & \checkmark & \xmark & \xmark & \xmark & \xmark & \checkmark \\
\textbf{BrainNav}~\cite{ling2025endowing} & \checkmark & \xmark & \xmark & \xmark & \checkmark & \checkmark \\
\bottomrule
\end{tabular}}
\end{table}

\subsubsection{Path Planning and Navigation for VLN Agents}
VLN agents must combine perception, reasoning, and planning to execute goal-directed navigation from natural-language instructions, as in Table~\ref{tab:vln_navigation_comparison}. LLMs often serve as the high-level planners in these systems.  
\textsc{NavVLM}~\cite{yin2024navigation} employs a VLM as the cognitive core, interpreting language goals and guiding exploration through semantic understanding of the environment.  
To enhance spatial reasoning, \textsc{SpatialCoT}~\cite{liu2025spatialcot} introduces bi-directional spatial coordinate alignment and Chain-of-Thought grounding, improving reasoning accuracy and interpretability.  

Addressing domain adaptation, \textsc{NavCoT}~\cite{lin2025navcot} uses parameter-efficient adaptation to enable self-guided navigation, generating coherent reasoning chains aligned with downstream planning. To reduce hallucinated plans, \textsc{FlexVLN}~\cite{zhang2025flexvln} validates LLM-generated guidance through an auxiliary MLLM, ensuring action feasibility.  
For long-horizon tasks, \textsc{NavA$^3$}~\cite{zhang2025nava} adopts a hierarchical framework: a reasoning VLM identifies target regions, and a pointing VLM performs fine-grained localization via spatial affordances.  

Mapping-based approaches further improve navigation. \textsc{TopV-Nav}~\cite{zhong2024topv} constructs adaptive top-view maps using visual prompts, providing structured spatial priors for reasoning. \textsc{BrainNav}~\cite{ling2025endowing} integrates dual maps (coordinate and topological) and dual orientations (relative and absolute), enabling real-time navigation with dynamic scene updates.  


\noindent \textit{\textbf{Insights \& Discussion.}} Recent methods enhance VLN agents by combining LLM-based planning with spatial grounding, domain adaptation, and hallucination mitigation. Structured spatial priors further support real-time reasoning. Future efforts should unify spatial perception and language reasoning for generalizable, low-supervision navigation.

\begin{table}[t]
\centering
\caption{Comparison of representative methods for Embodied Question Answering (EQA). \checkmark indicates the method supports or explicitly incorporates the feature.}
\label{tab:eqa_comparison}
\setlength{\tabcolsep}{4pt}
\resizebox{\linewidth}{!}{
\begin{tabular}{l|c|c|c|c|c}
\toprule
\textbf{Method} & \rotatebox{45}{Open-Vocab} & \rotatebox{45}{3D Scene Graph} & \rotatebox{45}{CoT Reasoning} & \rotatebox{45}{RL} & \rotatebox{45}{Modular Percept.} \\
\midrule
\textbf{Majumdar \etal~\cite{majumdar2024openeqa}} & \checkmark & \xmark & \xmark & \xmark & \xmark \\
\textbf{Tan \etal~\cite{tan2023knowledge}} & \xmark & \checkmark & \xmark & \xmark & \xmark \\
\textbf{Hao \etal~\cite{hao2024embosr}} & \xmark & \xmark & \checkmark & \xmark & \xmark \\
\textbf{Zhao \etal~\cite{zhao2025embodied}} & \xmark & \xmark & \checkmark & \checkmark & \checkmark \\
\bottomrule
\end{tabular}
}
\end{table}

\subsection{Multimodal Spatial Reasoning in Other Embodied Tasks}
\subsubsection{Embodied Question Answering (EQA)}
EQA, first proposed by Das \etal~\cite{das2018embodied}, has become a central benchmark in embodied AI and robotics. In this task, an agent receives a natural-language question—e.g., \textit{“Is there a sofa in the living room?”}—and must explore the environment, gather visual evidence, and provide an answer. The challenge lies in grounding language to spatial perception and reasoning.  
Majumdar \etal~\cite{majumdar2024openeqa} developed an open-vocabulary EQA dataset to evaluate foundation models, revealing that current systems struggle with spatial queries requiring object-level and scene-level understanding. To improve spatial reasoning, Tan \etal~\cite{tan2023knowledge} introduced a 3D scene graph as an external memory, enabling the model to retain and reason over spatial layouts across multiple turns, significantly improving multi-step QA efficiency. Hao \etal~\cite{hao2024embosr} advanced this direction by integrating Chain-of-Thought (CoT) reasoning within the Embosr framework, allowing structured spatial inference across complex 3D scenarios.  
Zhao \etal~\cite{zhao2025embodied} further decoupled perception and reasoning by assigning visual understanding to large-scale VLMs and using a lightweight language model, optimized via reinforcement learning, for reasoning. Incorporating a slow-thinking mechanism enhances depth and reliability in spatial reasoning.

\noindent \textit{\textbf{Insights \& Discussion.}} EQA task highlights the intricate interplay between language grounding, visual perception, and spatial reasoning in interactive environments. A key insight from recent advances is that bridging the gap between low-level visual inputs and high-level task understanding requires combining the strong perceptual capabilities of foundation models with explicit reasoning mechanisms, such as scene graphs, neural program synthesis, and chain-of-thought prompting. Future efforts may benefit from further aligning spatial representations with language semantics and enhancing the memory efficiency of agents in multi-turn reasoning settings.

\subsubsection{Embodied Grasping}
Robotic grasping in cluttered environments remains difficult due to occlusions and complex object interactions, demanding fine-grained spatial reasoning.  
\textsc{ThinkGrasp}~\cite{qian2024thinkgrasp} introduces goal-driven language prompts that help identify and prioritize obstructing objects, enabling grasp planning even for heavily occluded targets.  
\textsc{FreeGrasp}~\cite{jiao2025free} represents objects as discrete keypoints and overlays visual markers to enhance GPT-4o’s zero-shot spatial reasoning.  
\textsc{AffordGrasp}~\cite{tang2025affordgrasp} integrates GPT-4o for in-context affordance reasoning, predicting graspable parts and intended functions, which are grounded using VLPart and Grounded-SAM for part-conditioned optimization.  
Similarly, \textsc{UniDiffGrasp}~\cite{guo2025unidiffgrasp} leverages GPT-4o to infer target semantics and functional parts from user input, combining multi-stage segmentation and diffusion-based sampling for dual-arm grasp generation in complex scenes.

\noindent \textit{\textbf{Insights \& Discussion.}} Cluttered environments, frequent object occlusions, and the need to follow strict temporal and spatial action sequences constitute the primary challenges in embodied grasping tasks. In such settings, spatial reasoning plays a particularly critical role. Using visual observations effectively and appropriately integrating the reasoning capabilities of VLMs are key to addressing these challenges.

\subsubsection{Embodied World Models}

Embodied world models simulate the dynamics of physical environments, supporting policy learning, data-driven simulation, and long-horizon planning. However, models relying solely on 2D pixel observations often fail to capture accurate spatial relationships, leading to incomplete scene representations and weak depth or pose estimation. Structurally consistent scene generation is therefore crucial for effective spatial reasoning and world modeling.

\textsc{EVA}~\cite{chi2024eva} integrates a video generation model with a visual–language model, combining reasoning with high-quality video synthesis. \textsc{TesserAct}~\cite{zhen2025tesseract} simulates temporal evolution in 3D environments, enabling realistic interactions such as object manipulation and drawer opening while maintaining spatial–temporal consistency across RGB-DN sequences. More recently, \textsc{3DFlowAction}~\cite{zhi20253dflowaction} predicts object-level scene flow for manipulation and employs GPT-4o~\cite{hurst2024gpt} to verify task completion by aligning rendered final states with language descriptions, linking physical dynamics with semantic evaluation.

\noindent \textit{\textbf{Insights \& Discussion.}}  
Embodied world models form the foundation for large-scale simulation data used to train embodied agents. Ensuring geometric and spatial consistency in these generated environments is critical for supporting accurate spatial reasoning and realistic embodied intelligence.

    \begin{table*}[t!]
\centering
\caption{Comparison of recent multimodal spatial reasoning methods in video QA.}
\resizebox{\textwidth}{!}{
\begin{tabular}{c|c|c|c|c|c|c}
\toprule
\textbf{Year} & \textbf{Task} & \textbf{Dataset} & \textbf{Benchmark} & \textbf{Method} & \textbf{Spatial Components} & \textbf{Code} \\
\midrule
Arxiv 2024 & MCVQA,  OEVQA,  \etc & - & - & VideoLLaMA2~\cite{damonlpsg2024videollama2} & Convolution Connector & \href{https://github.com/DAMO-NLP-SG/VideoLLaMA2}{link} \\ \midrule
ACL 2024 & Long Video-QA & - & - & VideoINSTA~\cite{liao-etal-2024-videoinsta} & Content-based Reasoning & \href{https://github.com/mayhugotong/VideoINSTA}{link} \\  \midrule
Arxiv 2024 & ScanQA,  OpenEQA & - & - & Coarse Correspondence~\cite{liu2024coarse} & Lightweight tracking model & - \\ \midrule
Arxiv 2024 & Video-QA & VSI-Bench & VSI-Bench\cite{yang2024thinking} & - & - & \href{https://github.com/vision-x-nyu/thinking-in-space}{link} \\ 
\midrule
Arxiv 2025 & Video-QA & Video-R1  & - & Video-R1~\cite{video-r1} & GRPO & \href{https://github.com/tulerfeng/Video-R1}{link} \\ \midrule
Arxiv 2025 & Depth Estimation & - & - & AETHER~\cite{aether} & - & \href{https://github.com/OpenRobotLab/Aether}{link} \\ \midrule
Arxiv 2025 & RSTR & - & V-STaR~\cite{cheng2025v} & - & - & \href{https://github.com/V-STaR-Bench/V-STaR}{link} \\ \midrule
Arxiv 2025 & Video-QA & SpaceR-151k & - & SpaceR~\cite{ouyang2025spacer} & Task-Specific GRPO Training & \href{https://github.com/OuyangKun10/SpaceR}{link} \\ \midrule
Arxiv 2025 & Video-QA & Ego-ST Bench & Ego-ST Bench & ST-R1~\cite{ego-st} & Long-CoT and GRPO & \href{https://github.com/WPR001/Ego-ST}{link} \\
\bottomrule
\end{tabular}
}
\label{tab:videoqa_comparison}
\end{table*}

\section{Spatial Reasoning with Video and Audio}
\subsection{Spatial Reasoning with Video}
Video inherently captures more information about a scene than static images,  leading to significant research into the spatial reasoning capabilities of MLLMs. Extending the reasoning abilities from image-based tasks to video-based understanding opens exciting new possibilities. However,  accurately reasoning about spatial properties and establishing correspondences in dynamic,  temporal scenes remains a persistent challenge. As proposed by Spatial-R1~\cite{ouyang2025spacer},  seven critical spatial reasoning tasks are essential in this domain: object relative distance,  object size estimation,  room size estimation,  object relative direction,  object appearance order,  object absolute distance,  and object counting.

We systematically review this emerging area and summarize the key characteristics of the existing methods,  as shown in Tab.~\ref{tab:videoqa_comparison}. Recent work has explored specialized architectures and training strategies to enhance spatial reasoning capabilities in MLLMs. A representative example is Spatial-R1~\cite{ouyang2025spatial},  which proposes fine-tuning vision-language models with reinforcement signals grounded in spatial consistency. This training encourages the model to align outputs with the underlying 3D or 2D geometry implied by the video. 
SpaceR~\cite{ouyang2025spacer} further refines this approach by injecting positional tokens derived from visual object tracking,  enabling improved frame-to-frame localization. Other works introduce complementary strategies. R1-Zero-like training~\cite{vis100k} builds on reinforcement objectives to penalize spatial hallucinations and reward temporally stable spatial predictions. ST-Think~\cite{ego-st} introduces a dual-modality backbone that processes egocentric video using both motion and layout cues, enabling 4D (space-time) reasoning through transformer modules. Similarly,  Video-R1~\cite{video-r1} augments the visual encoder with spatial maps derived from frame-wise geometric analysis,  and uses spatial alignment loss to preserve inter-frame consistency. LLaVA-ST~\cite{st-align} and VideoINSTA~\cite{liao-etal-2024-videoinsta} adopt an orthogonal approach: they focus on instruction tuning with spatial-temporal prompts,  encouraging zero-shot understanding of video-level concepts like object permanence and navigational intent. These models rely on vision encoders (typically CLIP variants) that preserve spatial resolution via patch-wise tokenization. In Thinking in Space~\cite{yang2024thinking},  spatial memory is modeled explicitly through a recurrent memory cache,  allowing the LLM to recall visual states at earlier timestamps for long-horizon reasoning. A benchmark-centric perspective is introduced by V-STaR~\cite{cheng2025v},  which offers a suite of probing tasks to evaluate spatial reasoning across different axes: motion tracking,  occlusion recovery,  topological layout understanding,  and cross-frame object matching. Coarse Correspondence~\cite{liu2024coarse} complements this with a strategy that boosts spatial alignment across frames via coarse-to-fine token matching,  improving temporal coherence in reasoning chains. Lastly,  Aether~\cite{aether} proposes geometric-aware world modeling through unified token representations that encode both position and object identity,  enabling downstream LLMs to simulate spatial transitions with minimal hallucination.

\noindent \textit{\textbf{Insights \& Discussion.}} 
Recent progress in multimodal spatial reasoning demonstrates the growing capability of MLLMs to handle structured space-time understanding. However,  challenges remain: models often lose spatial detail due to token compression and lack mechanisms for robust spatial memory. Solutions such as marker-based overlays (as in MPDrive-style approaches) and coordinate-augmented prompts (as in LocVLM~\cite{ranasinghe2024learning}) provide partial remedies,  but fall short in generalizing across diverse video domains. Egocentric video in particular poses unique difficulties for multimodal spatial reasoning: distinguishing between agent motion and object motion requires grounded scene representations and persistent memory. While early efforts such as ST-Think and Thinking in Space offer promising architectures,  scalable and generalizable spatial world models remain an open research area.

\begin{table*}[ht!]
\centering
\caption{Comparison of recent multimodal spatial reasoning methods in audio.}
\resizebox{\textwidth}{!}{
\begin{tabular}{c|c|c|c|c|c}
\toprule
\textbf{Year} & 
 \textbf{Task} & \textbf{Benchmark} & \textbf{Method} & \textbf{Spatial Components} & \textbf{Code} \\
\midrule
NeurIPS 2023 & Audio-Visual Sound Localization and Detection & STARSS23~\cite{shimada2023starss23} & - & - & - \\ \midrule
ICML 2024 & Audio-QA & SpatialSoundQA~\cite{zheng2024bat} & BAT & Spatial Audio Encoder,  Curriculum Learning & \href{https://github.com/X-LANCE/SLAM-LLM/tree/main/examples/seld_spatialsoundqa}{link} \\ \midrule
ICML 2025 & Audio-QA &  AQAPHY & ACORN~\cite{wang2025teaching} &  Fundamental Physical Phenomena & - \\ \midrule
Arxiv 2025 & Audio-Visual-QA &  SAVVY-Bench & SAVVY~\cite{chen2025savvy} & Spatial Tracks and Global Map Construction & \href{https://huggingface.co/datasets/ZijunCui/SAVVY-Bench}{link} \\ 
\bottomrule
\end{tabular}
}
\label{tab:audio_comparison}
\end{table*}

\subsection{Spatial Reasoning with Audio}
\label{sec:audio}
Audio spatial reasoning is the process of interpreting spatial cues from sound,  such as direction of arrival,  source location,  and distance,  to infer the physical context of an auditory scene. While human listeners effortlessly localize and segregate sounds using binaural cues,  current multimodal large language models (MLLMs) have primarily focused on what is heard (the content) rather than where it is heard from~\cite{tang2024can}. This lack of spatial awareness limits applications such as audio-visual navigation and egocentric perception,  where an AI agent must infer where a sound originates to interact effectively with its environment. To bridge this gap,  recent research~\cite{yun2021pano,  tang2024can,  shimada2023starss23,  zheng2024bat,  wang2025teaching,  chen2025savvy} has begun to explore spatial reasoning capabilities by training large-scale multimodal models that learn from audio-only or audio-visual inputs.

\begin{figure}
    \centering
    \includegraphics[width=1\linewidth]{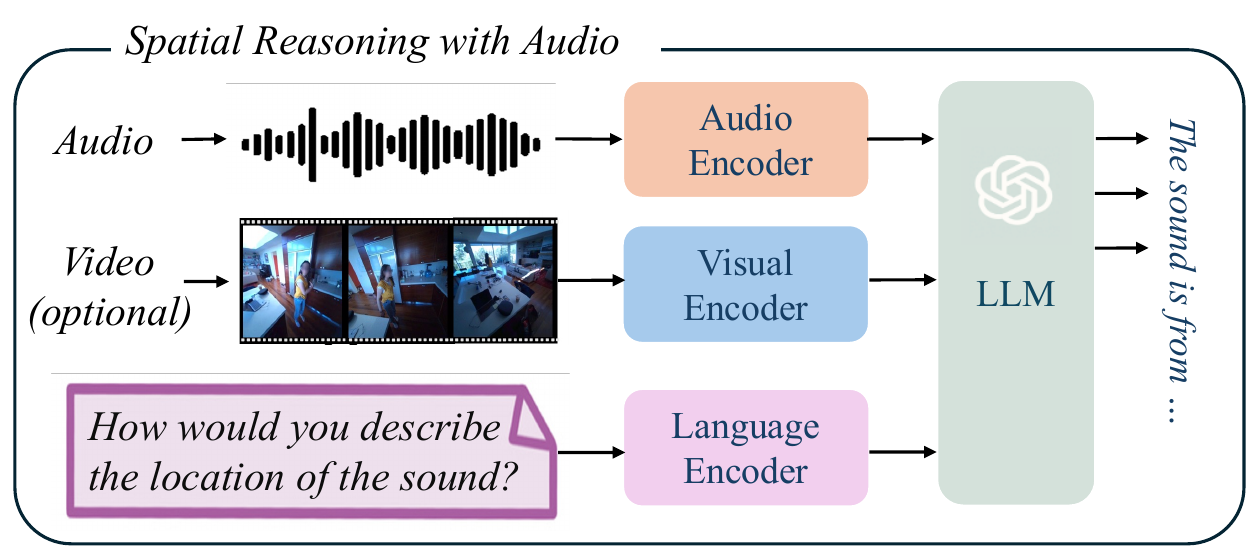}
    \caption{Spatial reasoning from audio \& video with MLLMs.}
    \label{fig:llmgaudio}
\end{figure}

We systematically review this emerging area and summarize the key characteristics of recently proposed methods,  as shown in Tab.~\ref{tab:audio_comparison}. STARSS23~\cite{shimada2023starss23} introduces an audio-visual sound event localization and detection (SELD) task,  along with the STARSS23 audio-visual dataset to support spatial reasoning for SELD. SpatialSoundQA~\cite{zheng2024bat} is the first large-scale benchmark focused on spatial audio question answering (Audio-QA). It includes over 21, 000 simulated binaural audio clips rendered in 3D environments,  accompanied by diverse questions involving directionality,  distance estimation,  and multi-source spatial reasoning. Architecturally,  the proposed BAT model combines a spatial audio encoder with a large language model (LLM) and employs curriculum learning to gradually enhance the model’s spatial reasoning capabilities.
ACORN~\cite{wang2025teaching} also addresses Audio-QA by introducing the AQAPHY benchmark. Technically,  it improves an LLM’s spatial reasoning by incorporating fundamental physical phenomena such as the Doppler effect,  multipath propagation,  and spatial relationships. More recently,  SAVVY~\cite{chen2025savvy} has emerged as a prominent testbed for spatial reasoning that integrates both audio and visual cues,  i.e.,  audio-visual question answering (Audio-Visual-QA). Specifically,  SAVVY presents SAVVY-Bench,  which evaluates 3D spatial reasoning in dynamic scenes with synchronized spatial audio,  and proposes to enhance spatial understanding by first extracting spatial tracks and then constructing a global spatial map. These benchmarks collectively advance standardized evaluation for audio spatial reasoning and enable quantitative comparison across MLLMs with varying degrees of spatial awareness.
It is worth noting that other Audio-QA and Audio-Visual-QA methods,  such as SARI~\cite{wen2025sari},  Meerkat~\cite{chowdhury2024meerkat},  and EchoInk-R1~\cite{xing2025echoink},  are not discussed here as they do not specifically address spatial reasoning.

\noindent \textit{\textbf{Insights \& Discussion.}} 
Despite recent progress,  significant challenges remain for robust audio spatial reasoning. Current models still struggle to generalize in open-world scenarios with multiple,  dynamic sound sources. These limitations are further compounded by the scarcity of large-scale,  high-quality spatial audio datasets with precise annotations,  which makes it difficult to train models that perform well outside of controlled or simulated environments. To bridge these gaps,  promising directions include the development of richer data collection pipelines,  such as real-world egocentric recordings or improved simulation techniques that better approximate real acoustic conditions.  In parallel,  more specialized model architectures are expected to emerge to effectively leverage these spatial cues. By addressing both data and modeling challenges, future systems may achieve human-like ``spatial hearing", reasoning not only about what is heard but also where it occurs within complex, dynamic scenes.

    \begin{figure*}[h!]
    \centering
    \includegraphics[width=\textwidth]{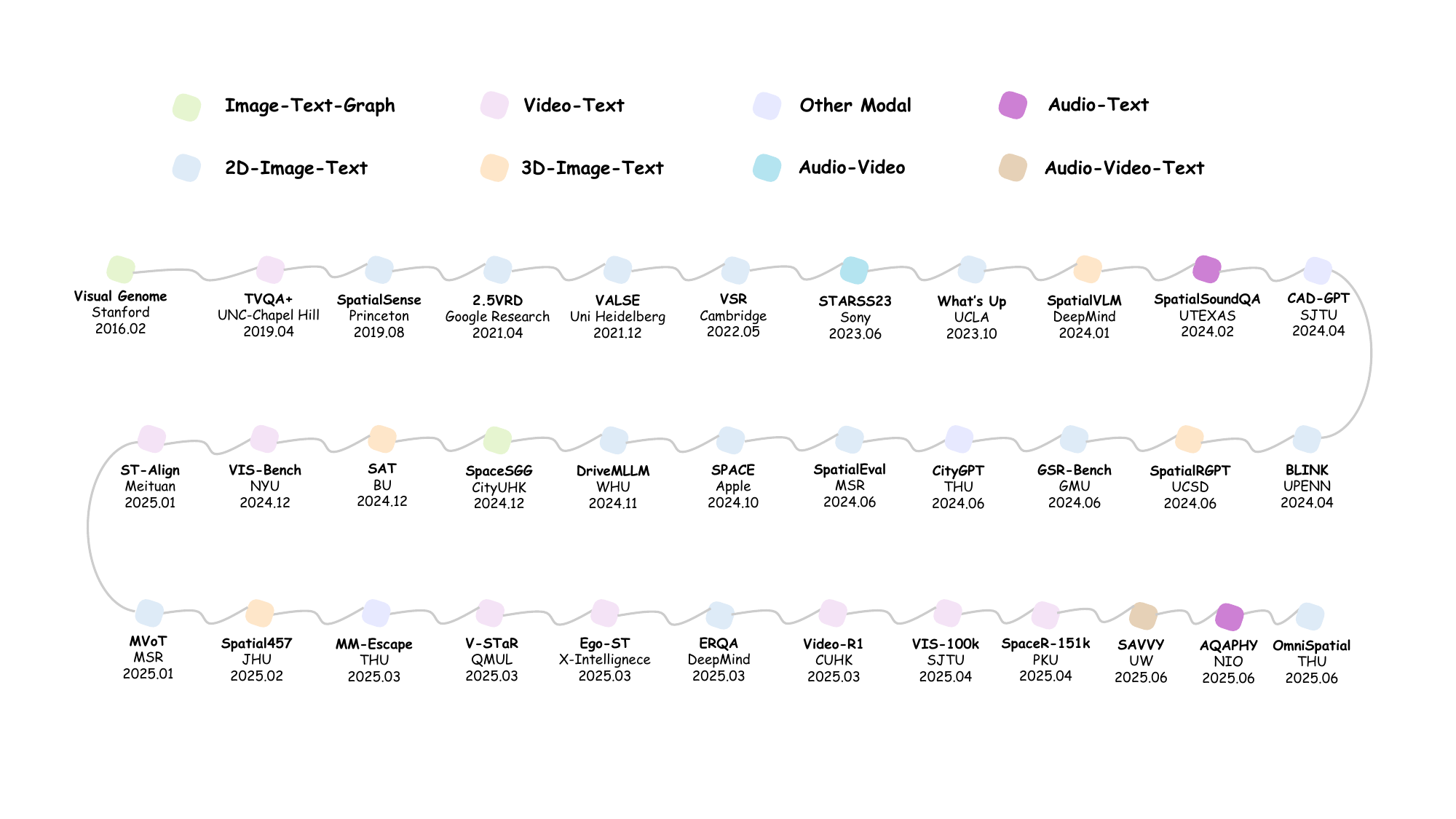}
    \vspace{-40pt}
    \caption{The chronological progression of multimodal spatial reasoning benchmarks. 
    Each colored marker represents a distinct benchmark, with hue variations indicating different 
    modality combinations (e.g., image-text-graph, audio-video). The timeline illustrates the 
    evolution of assessment methodologies and the increasing complexity of spatial reasoning 
    evaluation frameworks.}
    \label{fig:Bench_Timeline}
\end{figure*}

\section{Benchmarks}
\begin{table*}
\centering
\renewcommand{\tabcolsep}{16pt}
\resizebox{\linewidth}{!}{%
\begin{tabular}{lccccc}
\toprule
\textbf{Authors} & \textbf{Venue/Date} & \textbf{Paper Link} & \textbf{Code} & \textbf{Input Modality} \\
\midrule
Feng \etal & Arxiv 2025 (Mar) & \href{https://arxiv.org/pdf/2503.21776}{\faFilePdf} & \href{https://github.com/tulerfeng/Video-R1}{\faGithub} & Image-Text  \\
\hline
Imran Kabir \etal & Arxiv 2025 (Mar)  & \href{https://arxiv.org/abs/2503.12663}{\faFilePdf} & \href{https://github.com/Imran2205/LogicRAG}{\faGithub} & Video-Text  \\
\hline
Peiran Wu \etal & Arxiv 2025 (Mar) & \href{https://arxiv.org/abs/2503.12542}{\faFilePdf} & / & Video-Text  \\
\hline
Ziyue Wang \etal & Arxiv 2025 (Mar) & \href{https://arxiv.org/abs/2503.10042}{\faFilePdf} & \href{https://github.com/THUNLP-MT/EscapeCraft}{\faGithub} & Image-Text  \\
\hline
Jonathan Roberts \etal & Arxiv 2025 (Feb) & \href{https://arxiv.org/abs/2502.09696}{\faFilePdf} & \href{https://zerobench.github.io/}{\faGithub} & Image-Text  \\
\hline
Mingjie Xu \etal & WACV 2025 & \href{https://arxiv.org/abs/2412.06322}{\faFilePdf} & \href{https://github.com/Endlinc/LLaVA-SpaceSGG}{\faGithub} & Graph-Desc/QA/Conv  \\
\hline
Hongyu Li \etal & Arxiv 2025 (Jan) & \href{https://arxiv.org/abs/2501.08282}{\faFilePdf} & \href{https://github.com/appletea233/LLaVA-ST}{\faGithub} & Video-Text(QA)  \\
\hline
Yang \etal & CVPR 2025 & \href{https://arxiv.org/pdf/2412.14171}{\faFilePdf} & \href{https://github.com/vision-x-nyu/thinking-in-space}{\faGithub} & Video-Text(QA)  \\
\hline
Xingrui Wang \etal & CVPR 2025 & \href{https://arxiv.org/abs/2502.08636}{\faFilePdf} & \href{https://github.com/XingruiWang/Spatial457}{\faGithub} & Image-Text  \\
\hline
Liao \etal & Arxiv 2025 (Apl) & \href{https://arxiv.org/pdf/2504.00883}{\faFilePdf} & \href{https://github.com/zhijie-group/R1-Zero-VSI}{\faGithub} & Video-Text(QA)  \\
\hline
Chengzu Li \etal & Arxiv 2025 (Jan) & \href{https://arxiv.org/abs/2501.07542}{\faFilePdf} & / & Image-Text \\
\hline
Huanqia Cai \etal & Arxiv 2025 (Feb) & \href{https://arxiv.org/abs/2502.00698}{\faFilePdf} & \href{https://acechq.github.io/MMIQ-benchmark/}{\faGithub} & Image-Text \\
\hline
Siyu Wang \etal & AAAI 2025 & \href{https://ojs.aaai.org/index.php/AAAI/article/view/32849}{\faFilePdf} & \href{https://openiwin.github.io/CAD-GPT/}{\faGithub} & CAD-Text  \\
\hline
Navid Rajabi \etal & NIPS 2024 Workshop & \href{https://arxiv.org/abs/2406.13246}{\faFilePdf} & / & Image-Text(QA)  \\
\hline
Chonghao Sima \etal & ECCV 2024 & \href{https://arxiv.org/abs/2312.14150}{\faFilePdf} & \href{https://github.com/OpenDriveLab/DriveLM}{\faGithub} & Image/Graph-Text(QA)  \\
\hline
Ivan Majic \etal & GeoAI 2024 & \href{https://dl.acm.org/doi/abs/10.1145/3687123.3698293}{\faFilePdf} & \href{https://github.com/ivan-majic/llm_modality_reasoning}{\faGithub} & Image-Text  \\
\hline
Li Xuan \etal & IOTMMIM 24 & \href{https://dl.acm.org/doi/abs/10.1145/3698385.3699875}{\faFilePdf} & / & Image-Text  \\
\hline
Yew Ken Chia \etal & ACL 2024 & \href{https://arxiv.org/abs/2403.13315}{\faFilePdf} & \href{https://github.com/declare-lab/LLM-PuzzleTest}{\faGithub} & Image-Text  \\
\hline
Xiao Liu \etal & ACL 2022 & \href{https://arxiv.org/abs/2203.08075}{\faFilePdf} & \href{https://github.com/xxxiaol/spatial-commonsense}{\faGithub} & Image-Text  \\
\hline
Roshanak Mirzaee \etal & NAACL 2021 & \href{https://arxiv.org/abs/2104.05832}{\faFilePdf} & \href{https://github.com/HLR/SpartQA_generation}{\faGithub} & Text  \\
\hline
Yu-Chuan Su \etal & Arxiv 2021(Apr) & \href{https://arxiv.org/abs/2104.12727}{\faFilePdf} & \href{https://github.com/google-research-datasets/2.5vrd}{\faGithub} & Image-Text  \\
\hline
Letitia Parcalabescu \etal & ACL 2022 & \href{https://arxiv.org/abs/2112.07566}{\faFilePdf} & \href{https://github.com/Heidelberg-NLP/VALSE}{\faGithub} & Image-Text  \\
\hline
Liu \etal & TACL 2023 & \href{https://arxiv.org/pdf/2205.00363}{\faFilePdf} & \href{https://github.com/cambridgeltl/visual-spatial-reasoning}{\faGithub} & Image-Text  \\
\hline
Ramakrishnan \etal & Arxiv 2024 (Oct) & \href{https://arxiv.org/pdf/2410.06468}{\faFilePdf} & / & Image-Text  \\
\bottomrule
\end{tabular}}
\caption{General MLLM: Benchmarks and Datasets}
\end{table*}

Multimodal spatial reasoning enables AI systems to understand and infer spatial relationships within scenes by integrating information from multiple modalities, such as vision and language. Initially, benchmarks and datasets focused on simple scenes and basic spatial relations. However, as multimodal foundation models evolved, the focus shifted to more complex reasoning and cross-modal inference. Before these models, research was constrained to environments with basic spatial tasks, such as determining relative object positions in visual question answering (VQA). With the rise of powerful pre-trained models, new benchmarks were developed to address greater openness, richer complexity, and deeper reasoning capabilities. These efforts span domains like panoramic imagery, video, computer-aided design (CAD), and geographic information systems (GIS), advancing AI systems in scene understanding. Figure \ref{fig:Bench_Timeline} illustrates the development of multimodal spatial reasoning benchmarks. This section provides an overview of the evolution of datasets and benchmarks, highlighting key stages, modality types, and domain coverage, with a focus on those from the foundation model era.

\subsection{Early Multimodal Spatial Reasoning Benchmarks}

Before the advent of large-scale multimodal foundation models, early research in spatial reasoning relied heavily on datasets focused on natural images paired with textual descriptions. These datasets aimed to tackle basic spatial reasoning tasks, such as object localization and relationship detection.

A pivotal benchmark in this domain is the Visual Genome dataset~\cite{Genome}, which provides annotated images and graphs to depict spatial relationships between objects, facilitating image-text question answering tasks. Another significant contribution is SpatialSense~\cite{SpatialSense}, which contains a wide variety of spatial relationships, promoting tasks that involve misclassification-prone scenarios. Similarly, TVQA+~\cite{TVQA+} combines video clips with object detection annotations, requiring models to answer questions that involve both spatial and temporal reasoning. The 2.5VRD dataset~\cite{vrd2.5} focuses on fine-grained visual relationship detection using triplet annotations, capturing spatial relationships between objects.

Additionally, the VALSE benchmark~\cite{valse}, though not solely designed for spatial reasoning, includes rich annotations of spatial relationships and actions, providing an excellent resource for evaluating models' vision-language grounding capabilities. Further contributions, such as the VSR dataset~\cite{VSR}, define explicit spatial reasoning tasks, while datasets like COCO-Spatial in What'sUp~\cite{What'sUp} examine the limitations of pre-trained models on spatial reasoning. These early benchmarks, while focused on basic spatial cognition, set the foundation for more advanced tasks, sparking further developments in multimodal spatial reasoning for large-scale models.

\subsection{Image-Text Spatial Reasoning Benchmarks}

With the rise of large-scale multimodal foundation models (MLLMs), spatial reasoning tasks have expanded into various domains. This section discusses the evolution of 2D spatial reasoning benchmarks, categorizing them based on task objectives and methodologies.

\subsubsection{2D Spatial Reasoning Tasks}

2D spatial reasoning benchmarks evaluate models' ability to reason about spatial relationships in two-dimensional settings, focusing on tasks like navigation, object localization, and layout generation. A key trend is the integration of multimodal data, combining visual and textual information for enhanced reasoning. For example, DriveMLLM~\cite{drivemllm} annotates spatial relationships in driving scenarios using question-answer pairs, assessing navigation understanding. SpatialEval~\cite{worth} provides synthetic images with spatial tasks, such as Spatial-Map and Maze-Nav, testing relative object positioning in controlled settings. The SPACE benchmark~\cite{SPACE} offers both large-scale and small-scale tasks, from layout understanding to viewpoint transformations, evaluating models’ ability to handle diverse spatial challenges.

\subsubsection{Hybrid Approaches and Abstract Representations}

Some datasets explore abstract representations. VSR~\cite{VSR} provides annotations for spatial positional relationships, testing complex spatial reasoning. Datasets like COCO-Spatial~\cite{What'sUp} introduce spatial tasks that involve context, navigation, and dynamic reasoning. Other benchmarks, such as OmniSpatial~\cite{omnispatial} and GSR-Bench~\cite{gsr-bech}, enhance real-world relevance, offering comprehensive evaluations in areas like autonomous driving and robotics. OmniSpatial tests tasks like dynamic reasoning, traffic analysis, and geometric decomposition, reflecting real-world spatial complexities.

\subsubsection{Insights \& Discussion}

2D spatial reasoning datasets have evolved from simple image-text pairs to multi-task frameworks evaluating diverse reasoning abilities. Recent datasets emphasize multimodal data, combining visual and textual information for complex reasoning. Although synthetic data accelerates benchmarking, it faces challenges in generalization and real-world applicability. Future benchmarks should integrate dynamic real-world data and hybrid datasets combining synthetic and real data to better cover edge cases and enhance evaluation. These advancements will enable more capable models for autonomous navigation, robotics, and other complex applications.

\subsubsection{3D Spatial Reasoning Benchmarks}

The development of 3D spatial reasoning datasets has significantly advanced in recent years. Boyuan Chen \etal introduced the first 3D spatial reasoning dataset~\cite{spatialvlm}, incorporating depth-aware reasoning into multimodal systems. To evaluate MLLMs on dynamic spatial reasoning tasks, Arijit Ray \etal proposed the SAT dataset~\cite{SAT}, which includes simulated 3D scenes for training and real-world environments for testing. This dataset, using 3D scene simulations, improves model performance on dynamic spatial reasoning tasks through real-world evaluation.

To further address gaps in 3D spatial reasoning, An-Chieh Cheng \etal introduced SpatialRGPT-Bench~\cite{cheng2024spatialrgpt}, which generates 3D reasoning tasks grounded in 2D scenes. Their pipeline combines instance segmentation and depth estimation to construct tasks such as object size, height, and relative distance estimation using only 2D inputs. Additionally, Xingrui Wang \etal developed the Spatial457 dataset~\cite{spatial457} for 6D spatial reasoning, covering 3D localization, orientations, and multi-object relationships, further assessing the performance of MLLMs on these complex tasks.

\noindent \textit{\textbf{Insights \& Discussion.}} 
The introduction of 3D spatial reasoning benchmarks has brought significant advances, especially in data generation. Synthesis-driven annotation methods and automated 2D-to-3D conversion pipelines have alleviated annotation challenges. As tasks evolve, they have shifted from basic orientation and static perception to dynamic scene understanding and multi-perspective reasoning, increasing cognitive complexity. Furthermore, evaluation frameworks have transitioned from simulation-based training to real-world scenario validation, establishing closed-loop paradigms for performance assessment. Despite these advances, challenges remain, particularly in cross-modal alignment and adapting to dynamic scenes, highlighting the need for continued research in these areas.

\begin{table}[t!]
\captionsetup{font=small}
\caption{Performance comparison on Video--Text Spatial Reasoning Benchmarks (reported pairs only). Metrics follow the originals: \textsc{SpatialRGPT-Bench}—Success Rate; \textsc{BLINK}—Accuracy (spatial subset); \textsc{SpatialEval}—Accuracy (0–1); \textsc{DriveMLLM}—Zero-shot Score; \textsc{SAT}—Accuracy on Real/Synthetic.}
\label{tab:mllm_benchmarks}
\footnotesize
\setlength{\tabcolsep}{4pt}
\begin{tabularx}{\linewidth}{lllc}
\toprule
\textbf{Benchmark} & \textbf{Metric} & \textbf{Model} & \textbf{Value} \\
\midrule

\benchrow{\textsc{SpatialRGPT-Bench}~\cite{cheng2024spatialrgpt}}
 & Success Rate & LLaVA-v1.6-34B~\cite{Liu_2024_CVPR} & 43.98 \\
 & Success Rate & GPT-4V~\cite{gpt4} & \textbf{58.14} \\
\addlinespace[2pt]

\benchrow{\textsc{BLINK} (spatial)~\cite{blink}}
 & Acc. & LLaVA-v1.6-34B~\cite{Liu_2024_CVPR} & 76.22 \\
 & Acc. & InstructBLIP-Vicuna-7B~\cite{instructblip} & 55.24 \\
 & Acc. & InstructBLIP-Vicuna-13B~\cite{instructblip} & 64.34 \\
 & Acc. & Gemini-Pro~\cite{gemini} & 67.13 \\
 & Acc. & GPT-4V~\cite{gpt4} & 72.03 \\
 & Acc. & GPT-4o~\cite{gpt4} & \textbf{76.92} \\
\addlinespace[2pt]

\benchrow{\textsc{SpatialEval}~\cite{worth}}
 & Acc. (0--1) & LLaVA-v1.6-Mistral-7B~\cite{Liu_2024_CVPR} & 0.33 \\
 & Acc. (0--1) & LLaVA-v1.6-Vicuna-7B~\cite{Liu_2024_CVPR} & 0.24 \\
 & Acc. (0--1) & LLaVA-v1.6-Vicuna-13B~\cite{Liu_2024_CVPR} & 0.38 \\
 & Acc. (0--1) & LLaVA-v1.6-34B~\cite{Liu_2024_CVPR} & 0.42 \\
 & Acc. (0--1) & InstructBLIP-Vicuna-7B~\cite{instructblip} & 0.24 \\
 & Acc. (0--1) & InstructBLIP-Vicuna-13B~\cite{instructblip} & 0.27 \\
 & Acc. (0--1) & Gemini-Pro~\cite{gemini} & 0.687 \\
 & Acc. (0--1) & GPT-4V~\cite{gpt4} & \textbf{0.924} \\
\addlinespace[2pt]

\benchrow{\textsc{DriveMLLM}~\cite{drivemllm}}
 & Score (ZS) & LLaVA-v1.6-Mistral-7B~\cite{Liu_2024_CVPR} & 38.20 \\
 & Score (ZS) & LLaVA-v1.6-Vicuna-7B~\cite{Liu_2024_CVPR} & 38.20 \\
 & Score (ZS) & LLaVA-v1.6-Vicuna-13B~\cite{Liu_2024_CVPR} & 38.20 \\
 & Score (ZS) & LLaVA-ov-7B~\cite{llava-ov} & 22.29 \\
 & Score (ZS) & LLaVA-ov-72B~\cite{llava-ov} & 21.10 \\
 & Score (ZS) & Qwen2-VL-7B~\cite{qwen2vl} & 21.17 \\
 & Score (ZS) & Qwen2-VL-72B~\cite{qwen2vl} & 20.11 \\
 & Score (ZS) & Qwen-VL~\cite{qwenvl} & 36.50 \\
 & Score (ZS) & mPLUG-Owl2~\cite{mplug-owl2} & 33.90 \\
 & Score (ZS) & InstructBLIP-Vicuna-7B~\cite{instructblip} & 42.80 \\
 & Score (ZS) & InstructBLIP-Vicuna-13B~\cite{instructblip} & 42.80 \\
 & Score (ZS) & Gemini-1.5-flash~\cite{gemini1.5} & \textbf{54.03} \\
 & Score (ZS) & Gemini-Pro~\cite{gemini} & 40.10 \\
 & Score (ZS) & GPT-4V~\cite{gpt4} & 51.70 \\
 & Score (ZS) & GPT-4o~\cite{gpt4} & 25.63 \\
\addlinespace[2pt]

\benchrow{\textsc{SAT}~\cite{SAT}}
 & Acc. (Real)      & Gemini-1.5-flash~\cite{gemini1.5} & 57.60 \\
 & Acc. (Synthetic) & Gemini-1.5-flash~\cite{gemini1.5} & 50.00 \\
 & Acc. (Real)      & Gemini-1.5-Pro~\cite{gemini1.5} & \textbf{64.80} \\
 & Acc. (Synthetic) & Gemini-1.5-Pro~\cite{gemini1.5} & 49.90 \\
 & Acc. (Real)      & GPT-4V~\cite{gpt4} & 50.70 \\
 & Acc. (Synthetic) & GPT-4V~\cite{gpt4} & 44.80 \\
 & Acc. (Real)      & GPT-4o~\cite{gpt4} & 57.50 \\
 & Acc. (Synthetic) & GPT-4o~\cite{gpt4} & 49.40 \\
\bottomrule
\end{tabularx}
\vspace{-0.5em}
\end{table}

\subsection{Video-Text Spatial Reasoning Benchmarks}

Recent advancements in video-text spatial reasoning have led to the development of diverse benchmarks aimed at systematically evaluating spatial understanding capabilities. These benchmarks have evolved from fundamental perceptual tasks to more complex spatiotemporal tasks. Current benchmarks increasingly emphasize the integration of temporal and spatial cues, leveraging both synthetic and annotated data to support model training and evaluation. The following sections provide a detailed overview of these benchmarks, categorized by task type and complexity, highlighting their contributions in vedio-text spatial reasoning.

\subsubsection{Fundamental Spatial Perception Tasks}

Benchmarks in this category evaluate core spatial perception skills such as object counting, relative direction, and distance estimation.  
\textsc{VIS-100K}~\cite{vis100k} introduces 100,000 video–question–answer pairs spanning six spatial reasoning tasks—object count, relative/absolute distance, relative direction, object size, and room size. Fine-tuning MLLMs on this dataset demonstrates that the GRPO reinforcement algorithm effectively enhances spatial reasoning performance.  
\textsc{VIS-Bench}~\cite{yang2024thinking} further examines how MLLMs memorize and reason about spatial layouts. Built from 288 annotated indoor videos, it includes 5,000 QA pairs covering eight tasks such as distance, direction, path planning, and order of appearance, offering a detailed analysis of spatial understanding.  
\textsc{SpaceR-151K}~\cite{ouyang2025spacer} expands this scope with 151K samples, including 91K spatial QA pairs and 60K general video understanding examples. Each task incorporates precise spatial metadata (e.g., bounding boxes, temporal indices) and 10×10 grid maps encoding object distributions. Rigorous quality control ensures balanced, unambiguous data, establishing a new large-scale benchmark for spatial reasoning in multimodal systems.

\subsubsection{Advanced Spatiotemporal Reasoning Tasks}

These benchmarks extend spatial reasoning to dynamic tasks such as path planning and cross-modal coordination, emphasizing temporal consistency and causal reasoning.  
\textsc{ST-Align}~\cite{st-align} establishes a unified framework for fine-grained spatiotemporal reasoning with three tasks: Spatial-Temporal Video Grounding (STVG), Event Localization and Captioning (ELC), and Spatial Video Grounding (SVG). It jointly evaluates spatial and temporal localization, advancing beyond datasets focused on isolated spatial or temporal cues.  
\textsc{Ego-ST}~\cite{ego-st} addresses the overlooked role of temporal dynamics by introducing reverse egocentric reasoning. Comprising over 5,000 QA pairs across four tasks—route description, directional change, landmark transition, and action shift—it systematically evaluates how MLLMs integrate dynamic spatial cues and temporal order.  
\textsc{V-STaR}~\cite{cheng2025v} targets the gap between object-centric and temporal reasoning. Its core Reverse Spatio-Temporal Reasoning (RSTR) task links “What→When→Where” and “What→Where→When” chains to assess logical consistency, using the Logarithmic Geometric Mean (LGM) metric to jointly measure accuracy, temporal IoU, and spatial IoU. It establishes the first standardized benchmark for comprehensive spatiotemporal reasoning in Video-LLMs. Overall, these datasets advance spatial intelligence evaluation from static spatial perception to dynamic, temporally grounded reasoning—crucial for realistic embodied and video understanding.

\subsubsection{Mixed-Task Benchmarking}

This class of evaluation benchmarks incorporates diverse data sources and tasks of varying difficulty levels to provide a comprehensive assessment of model capabilities. Due to the scarcity of high-quality video reasoning data, current MLLMs exhibit limited spatial reasoning capabilities in video contexts. To address this issue, Feng \etal introduced two datasets: Video-R1-COT-165k and Video-R1-260k\cite{video-r1}. The former contains CoT annotated samples generated from both image and video inputs, serving as a cold-start dataset for supervised fine-tuning. The latter is designed for reinforcement learning training, comprising a mix of image and video data to enable models to acquire general reasoning skills from static images and transfer them to dynamic video contexts through a hybrid training strategy. Although only about 8\% of the samples in these datasets involve explicit spatial reasoning tasks, the inclusion of complete CoT annotations offers valuable resources for advancing research on spatial reasoning in video-based settings.

\noindent \textit{\textbf{Insights \& Discussion.}}
Current visual-spatial reasoning benchmarks are advancing from static attribute recognition toward dynamic spatiotemporal coupling, demanding progressively higher spatial cognitive capabilities from models; however, they remain constrained by limitations including prohibitive annotation costs restricting dataset scalability, inconsistent quality in semi-automated multimodal LLM-generated annotations, and overly homogeneous templated data that inadequately fosters profound spatial cognition—necessitating a paradigm shift from isolated data curation to synergistic algorithm-data co-design, from single-modality datasets to multi-source hybrid data frameworks, and from superficial pattern matching to causal inference incorporating physical constraints like gravitational collision dynamics.

\subsection{Other Modal Benchmarks}

Additional multimodal benchmarks extend spatial reasoning beyond vision–language inputs.  
For audio–visual spatial reasoning, related datasets and evaluation protocols are detailed in Section~\ref{sec:audio}.  
\textsc{CityInstruction} and \textsc{CityEval}, released with \textsc{CityGPT}~\cite{citygpt}, evaluate spatial reasoning, navigation, and path generation in realistic urban scenes.  
\textsc{CAD-GPT}~\cite{cad-gpt} introduces a dataset pairing natural language descriptions and single-view images with CAD modeling sequences, enabling multimodal 3D model synthesis and benchmarking.  
\textsc{SGG}~\cite{sgg} provides structured scene graphs and 3D point clouds fused with LLM-generated question–answer dialogues, supporting open-vocabulary spatial reasoning across complex visual layouts.  
Finally, \textsc{MM-Escape}~\cite{mm-escape} simulates an interactive escape-room environment where models must perform sequential spatial reasoning and actions to exit, offering a novel framework for evaluating goal-driven reasoning in dynamic scenes.

\noindent \textit{\textbf{Insights \& Discussion.}} Contemporary multimodal spatial reasoning datasets exhibit a tripartite evolution—progressing from scene-driven construction to task sophistication and evaluation closure—where real-world task demands grow increasingly complex, modalities diversify, and spatial reasoning advances beyond basic directional perception toward causal spatial inference chains. Nevertheless, persistent gaps remain in establishing a unified framework that ensures physical plausibility, enables action verifiability, and maintains cost-effective data curation, indicating considerable scope for advancement in multimodal spatial reasoning data infrastructure.
    \section{Challenges and Future Directions}

\noindent \textbf{Multimodal Spatial Reasoning in Egocentric Vision.}
While existing research on spatial reasoning in MLLMs primarily focuses on third-person perspectives, there is a growing need to explore egocentric vision, where spatial reasoning must occur from the agent's first-person viewpoint~\cite{plizzari2025omnia, li2025egocross,wen2025ai}. This shift introduces unique challenges, such as the agent's movement, limited field of view, and the temporally evolving nature of the environment. In egocentric vision, spatial reasoning must account for dynamic changes in both the agent’s position and the environment. Future research should focus on developing MLLMs capable of understanding object relationships from shifting viewpoints, inferring navigation intent, and reasoning about interaction affordances. A promising direction lies in creating models that can more effectively simulate and understand embodied behaviors, leading to more grounded, real-world intelligence. 

\noindent \textbf{Multimodal Spatial Reasoning in 3D Vision.}
Despite progress, current 3D MLLMs face challenges in scalability and interpretability due to the inherent complexity of 3D data. Additionally, the scarcity of large-scale annotated 3D datasets constrains the development of robust models. To address these challenges, future research should focus on the development of unified and efficient 3D representations that are both interpretable and scalable. Furthermore, training strategies that do not rely on large-scale annotated datasets, such as leveraging synthetic data, could offer valuable insights. By exploring the integration of symbolic reasoning into the 3D domain, researchers can ensure better handling of spatial relationships and improve model performance across unseen environments. A key goal should be creating frameworks that combine efficient 3D learning with strong temporal and causal reasoning capabilities to model dynamic spatial environments.

\noindent \textbf{Multimodal Spatial Reasoning in Embodied AI.}
Current methods for spatial reasoning in embodied AI often struggle to generalize to novel environments and are prone to spurious or hallucinated spatial inferences. Explicit reasoning modules, while improving inter-pretability, tend to increase inference overhead and still fall short in maintaining long-term spatial consistency. To advance this field, future research must focus on closer integration between perception and reasoning, ensuring that spatial models maintain both geometric fidelity and temporal consistency. Additionally, creating world models that combine sensory inputs (e.g., visual, auditory, tactile) with structured scene representations could allow for more robust spatial reasoning in dynamic environments. Scalable training strategies that incorporate symbolic and structured reasoning, along with the ability to perform causal inference over time, will be crucial in achieving long-term success in this area.

\noindent \textbf{Multimodal Spatial Reasoning with Novel Sensors.}
Emerging sensor technologies such as omnidirectional cameras~\cite{zhang2025omnidirectional,zhang2025towards,dongfang2025multimodal}, event cameras, LiDAR, thermal, and radar sensors offer complementary spatial information under challenging conditions like adverse lighting, weather, and high-speed motion~\cite{liao2025benchmarking}. However, these sensors introduce new challenges, including equirectangular distortions, orientation ambiguities, sparse and asynchronous data, and noise in radar and thermal signals. MLLMs, which are typically optimized for perspective RGB images, must evolve to effectively integrate and process these diverse modalities. Future research should focus on developing methods for fusing these heterogeneous sensor data into a unified spatial representation~\cite{zheng2024centering}, improving both the accuracy and robustness of spatial reasoning. By incorporating causal and temporal reasoning capabilities into sensor fusion, models can better handle dynamic environments and make more informed, context-aware decisions~\cite{zheng2025reducing}. Moreover, training strategies that leverage both synthetic and real-world sensor data could enhance model generalization across different sensor modalities.

\noindent \textbf{Multimodal Spatial Reasoning Benchmarks.}
Existing benchmarks are limited in their scope, often suffering from issues such as orientation under-specification, narrow modality coverage, and restricted interaction. To address these limitations, future work should focus on developing more comprehensive benchmarks that span a wider range of modalities and interaction settings. This includes constructing benchmarks that synchronize vision, depth, point clouds, panoramic views, spatial audio, inertial signals, and topological maps, all within a unified coordinate frame with explicit orientation and reference-frame labels. Future benchmarks should also focus on evaluating MLLMs' ability to perform tasks such as reference, navigation, inspection, and question answering in diverse environments. The development of interpretable evaluation frameworks that can assess both reasoning quality and spatial accuracy, while providing clear guidance for model improvement, will be essential. Additionally, incorporating symbolic reasoning into these benchmarks could allow for the assessment of structured spatial knowledge and enable better handling of complex real-world tasks.

    \section{Conclusion}
Large multimodal reasoning models have gradually emerged as a promising and critical solution toward achieving spatial reasoning capabilities.
In this paper, we focus on the intersection of spatial reasoning and MLLMs. 
Firstly, based on general spatial reasoning tasks, we systematically review and analyze the existing research from four perspectives: test-time scaling, post-training, model design, and explainability. 
We then extend the discussion to 3D vision tasks, including 3D visual grounding, 3D scene reasoning and question answering, and 3D generation. 
Beyond these fundamental tasks, we further explore spatial reasoning in embodied AI, providing reviews and discussions on vision-language navigation, embodied question answering, and related areas.
Moreover, spatial reasoning tasks involving emerging modalities such as video and audio are also summarized, which are challenging but crucial to building a comprehensive human-like spatial reasoning system.
In addition to methodological aspects, we provide a comprehensive overview of datasets and benchmarks for multimodal spatial reasoning, which constitute the indispensable support for advancements in this field. 
Through this systematic survey, we aim to establish a solid knowledge foundation and offer new insights to this field-paving the way toward intelligent and reliable multimodal spatial reasoning systems in the era of large models.

    \bibliographystyle{IEEEtran}
    \bibliography{ref}
\end{document}